\documentclass[conference]{IEEEtran}
\usepackage[a4paper,left=3cm,right=2cm,top=2.5cm,bottom=2.5cm]{geometry}
\usepackage{cite}
\usepackage[pdftex]{graphicx}
\usepackage[cmex10]{amsmath}
\usepackage{amssymb}

\usepackage{times}
\usepackage{array}
\usepackage[tight,footnotesize]{subfigure}
\usepackage[font=footnotesize]{subfig}
\usepackage{graphicx}
\usepackage{mathtools}
\usepackage{amsmath}

\def\intersect{{\cap}}

\usepackage{alltt}
\usepackage{shadethm}
\newshadetheorem{comment}{Comment}

\def\boxx{{\vcenter{\vbox{\hrule height.3pt
          \hbox{\vrule width.3pt height6pt
          \kern6pt\vrule width.3pt}\hrule height.3pt}}\;}}

\def\impos{{\;\vcenter{\hbox{\rule{5mm}{0.2mm}}} \vcenter{\hbox{\rule{1.5mm}{1.5mm}}} \;}}

\def\lrarrow{\leftrightarrow \kern-8pt \rightarrow}

\def\2{\frac{1}{2}}

\def\beq{\begin{eqnarray}}
\def\eeq{\end{eqnarray}}
\def\2{\frac{1}{2}}

\newtheorem{law}{Law}

\def\lrarrow{\leftrightarrow \kern-8pt \rightarrow}

\def\frightarrow{\rightarrow \kern-11pt /~~}
\def\reducesto{\simeq \kern -3pt >}

\usepackage{fancyhdr}					
\begin{document}
\newcommand{\strust}[1]{\stackrel{\tau:#1}{\longrightarrow}}
\newcommand{\trust}[1]{\stackrel{#1}{{\rm\bf ~Trusts~}}}
\newcommand{\promise}[1]{\xrightarrow{#1}}
\newcommand{\revpromise}[1]{\xleftarrow{#1} }
\newcommand{\assoc}[1]{{\xrightharpoondown{#1}} }
\newcommand{\rassoc}[1]{{\xleftharpoondown{#1}} }
\newcommand{\imposition}[1]{\stackrel{#1}{\impos}}
\newcommand{\scopepromise}[2]{\xrightarrow[#2]{#1}}
\newcommand{\handshake}[1]{\xleftrightarrow{#1} \kern-8pt \xrightarrow{} }
\newcommand{\cpromise}[1]{\stackrel{#1}{\frightarrow}}
\newcommand{\policy}{\stackrel{P}{\equiv}}
\newcommand{\field}[1]{\mathbf{#1}}
\newcommand{\bundle}[1]{\stackrel{#1}{\Longrightarrow}}

\title{A Spacetime Approach to Generalized Cognitive Reasoning in Multi-scale Learning\\~\\\small Semantics and knowledge representation from observation, measurement, and learning.}

\author{Mark Burgess\\~\\\small August 2016\\Revised January-February 2017\\Appendix examples added July 2017}
\maketitle
\IEEEpeerreviewmaketitle

\renewcommand{\arraystretch}{1.4}

\begin{abstract}
  In modern machine learning, pattern recognition replaces realtime
  semantic reasoning. The mapping from input to output is learned with
  fixed semantics by training outcomes deliberately. This is an
  expensive and static approach which depends heavily on the
  availability of a very particular kind of prior training data to
  make inferences in a single step.  Conventional semantic network
  approaches, on the other hand, base multi-step reasoning on modal
  logics and handcrafted ontologies, which are {\em ad hoc}, expensive
  to construct, and fragile to inconsistency.  Both approaches may be
  enhanced by a hybrid approach, which completely separates reasoning
  from pattern recognition. In this report, a quasi-linguistic
  approach to knowledge representation is discussed, motivated by
  spacetime structure. Tokenized patterns from diverse sources are
  integrated to build a lightly constrained and approximately
  scale-free network.  This is then be parsed with very simple
  recursive algorithms to generate `brainstorming' sets of reasoned knowledge.
\end{abstract}



\section{Introduction} 

Reasoning has long been associated with formal logic in computer
science, but it may be argued that reasoning is only a subset of a
wider class of narratives, which may be told by joining together
assertions.  Logic's goal is to maximize the certainty of a narrative
derived from prior concepts, by transforming them according to a
constrained and largely deterministic set of rules, in which each step
selects a unique possibility.  However, it is only of value in a very
limited set of circumstances.  In other cases, where unique certainty
is not available or practical, a more expansive kind of
`brainstorming' is needed for problem solving: one that allows
multiple possibilities to remain open for selection at a later time.

Brainstorming is the first stage of a process of reasoning by
`whittling'. The idea is to create a large hypothesis set, for subsequent
reduction into something more focused. The whittling is often emergent
and iterative.  Reasoning, in this interpretation, is a cognitive
process, which allows an observer to integrate experiences, perhaps
not evidentially linked, or even provably true, in order to form a set
of speculative hypotheses. By process of contextual elimination, this
converges, iteratively, to a much smaller set, or a final answer, using
a variety of criteria from freshness to relevance and importance.  In
this dynamical approach, criteria like semantic and dynamic {\em stability} of
the stories are now more important than logical idealizations like
`truth' or  ad hoc determinations of `correctness'\cite{certainty}.

Narration, or storytelling, is an underestimated aspect of intelligent
behaviour\cite{burgessaims2009,burgesskm,stories}. It is how learning
individuals explain things in terms of prior experience, and make
sense of the world. To form a story, we have to know the difference
between correlation and causation: correlations do not have direction,
i.e. no arrow of progression, which may be used to accumulate
narrative storyline. Hence, correlation can only offer short-range
explanations, with uncertainty that grows with the number of claims
for similarity. Combining directed associations into
a reasoned argument is a different problem altogether, one that
requires the propagation of semantics from step to step.

How we combine concepts into stories for small sets is one thing;
applying expansive reasoning to systems with vast numbers of sensors,
sporting very different characteristics, is a whole new challenge.
Imagine a scenario like the monitoring of a massive array of smart
connected systems, e.g. an Internet of Things, smart buildings and
cities, etc.  Data of very different kinds are generated at many
different scales, and from a vast number of different sources.  How
could we begin to interpret phenomena across different scales?  What
concepts do we need?  The challenge of making sense of such data, at
scale, urgently calls for a broader view of artificial reasoning,
based on improved semantic labelling of collected data.  This is how
we might scale the meaningful interpretation of data.  But it is not
simply a case of collecting more and more data.  The monitoring of
sensory data would be incomplete without the ability to reason about
observations.  Reasoning requires the definition of a set of concepts,
tied to operational goals.

This work summarizes a model of invariant storytelling, based on
spacetime inference\cite{spacetime1,spacetime2,spacetime3}. The
approach reduces the computational complexity for story inference by
orders of magnitude. It describes how semantic associations may be
collected from arbitrary sensors and inputs, and how a knowledge
representation can learn context-dependent interpretations of those
inputs, in order to later generate explanatory narrative
automatically. It summarizes the practical and implementational
aspects of the promise theoretic notion of `semantic spacetime', and
implements a number of experiments based on the model.

Promise theory\cite{promisebook} motivates a graph theoretical
approach to semantic scaling\cite{spacetime2}, which is based on the
observation that cognitive relationships ultimately derive from
elementary spacetime relationships. A set of recursive structures
leads to a partially deterministic set of connected paths, each of
which represents reasoned `stories', with explanatory content.  In
this work, I show how to apply these structures, tentatively, to simple cases.

\section{The semantic spacetime (promise) model}

\begin{figure}[ht]
\begin{center}
\includegraphics[width=6.5cm]{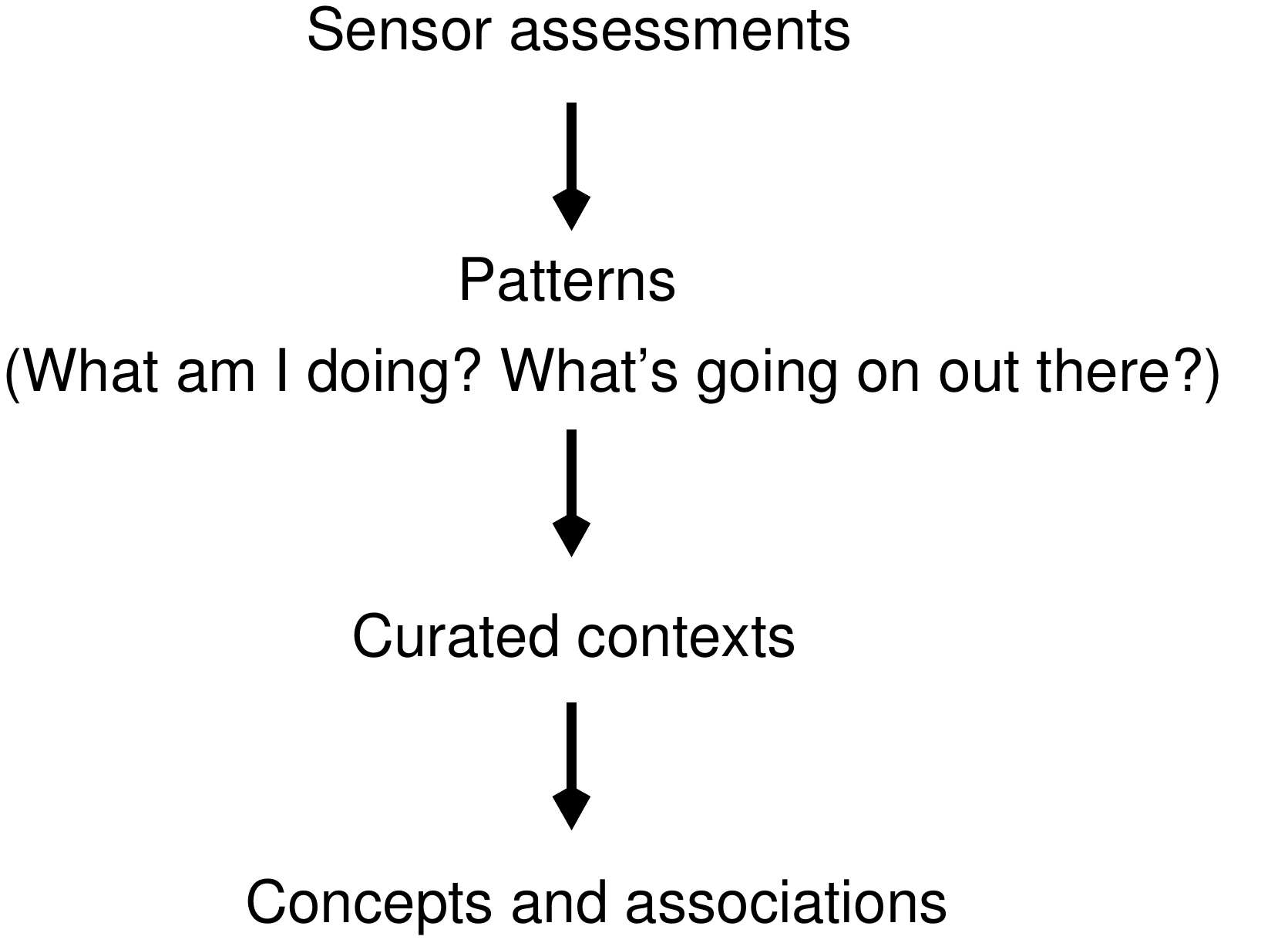}
\caption{\small From fast to slow, concepts are aggregated from
  co-activations over many scales. At each scale, meaning comes from
  the way we assign names to the aggregates.\label{scaleassoc}}
\end{center}
\end{figure}

Promise theory frames elementary questions about intent: how it scales
by cooperative interaction into extensive spatial networks, and how its
properties may express spacetime-like variations to mirror a
representation of environment.  Based on the considerations in
\cite{spacetime3}, we expect a number of information processing
stages in a cognitive learning system (figure \ref{scaleassoc}):
\begin{enumerate}
\item A sensory apparatus for inputting exterior information.
\item A stage that replaces sensory patterns with semantic tokens, i.e. {\em naming}.
\item A stage that associates semantics i.e. meaning to the named tokens, by associating
them as clusters called {\em concepts}, associated in predictable and {\em qualifying} ways.
\item An introspective story-generating stage, which is able to
  `think' about or activate interior concepts, and feed the resulting
  stream of consciousness back into the post-tokenization stages (3
  onwards), along side sensory inputs.
\end{enumerate}
Although we are not used to thinking of systems (physical or software) explained
as narratives, any organized arrangement may be understood in terms of stories,
so our goal is completely general.
Most discussions of so-called Artificial Intelligence (AI)
discuss stages 1 and 2, this work deals mainly with stages 3 and
onwards, which I believe are orthogonal. These latter stages have much
in common with linguistics\cite{spacetime3}; indeed, once one can map
invariant patterns into a finite alphabet of tokens, every reasoning
problem maps to a linguistic problem.

A simple prototype system, to explore the underlying principles, has
been reconstructed from the earlier cognitive approaches applied to
pervasive computing management in CFEngine\cite{burgessC1,lisa98283},
and further modernized and extended in the Cellibrium project \cite{cellibrium}.

\section{Separation of learning scales}

\begin{figure}[ht]
\begin{center}
\includegraphics[width=6.5cm]{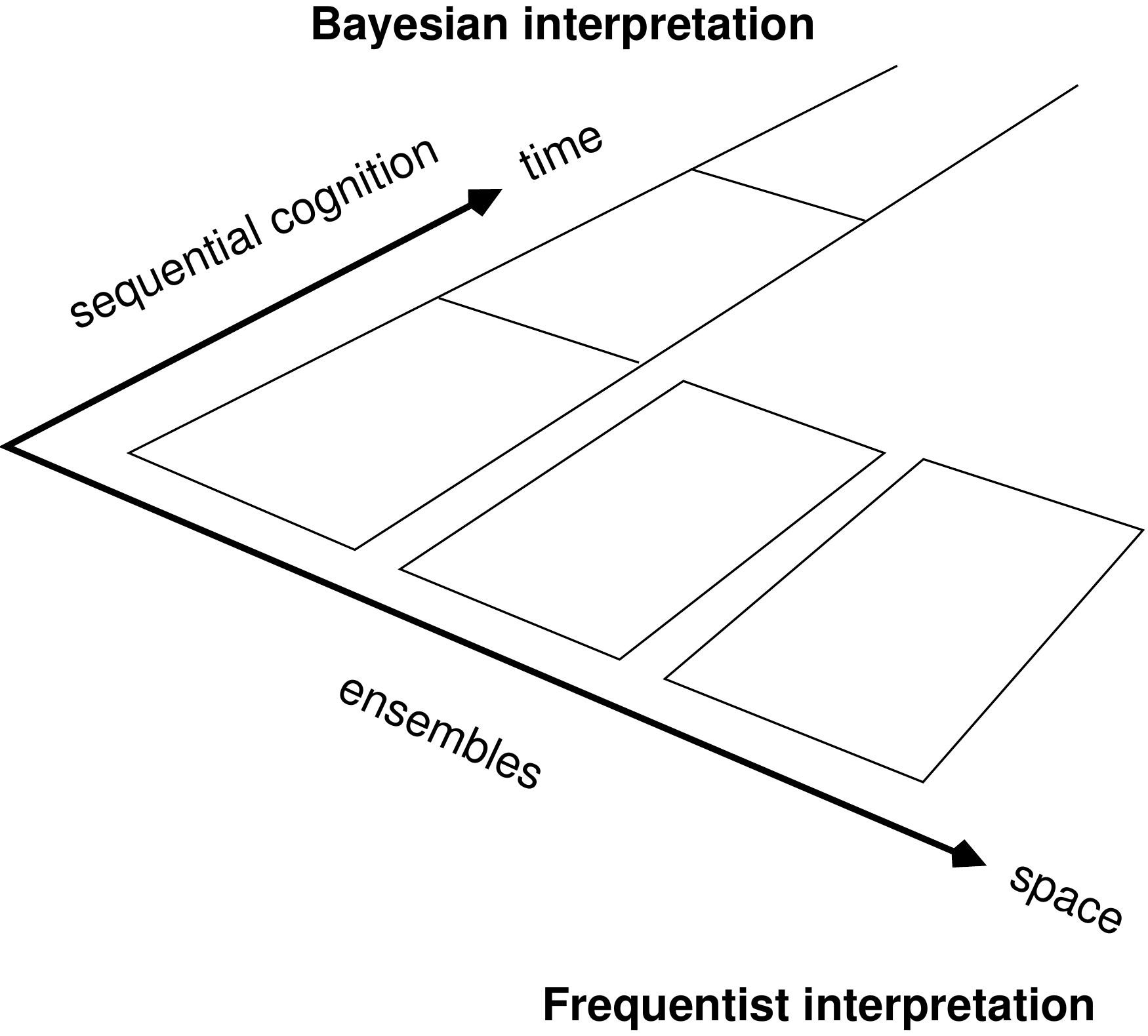}
\caption{\small We coarse grain space and time into regions of
  partially indistinguishable equivalence, e.g. different experimental
  `trials'. Concurrence (in the same temporal grain), and coincidence
  (same spatial grain), define the boundaries for our cognitive
  experience. In cognition, every new timelike frame is a new
  experiment, and we must use learning to even out experimental
  inequivalence. Our (un)certainty is related to our choice of
  scaling or granularity.\label{conchunks}}
\end{center}
\end{figure}

Data from only a single sensory episode, localized in time or space,
are of limited value, and do not typically lead to claims of deep
knowledge. It is by revisiting experiences, i.e. by the iterative
process of observing and learning that we build trust in a stable
`invariant' representation of knowledge (see figure \ref{conchunks}).

Learning happens over a hierarchy of timescales\cite{spacetime3}.
Fundamental concepts are evolved by stabilizing `genetic' adaptations
through long term selection in a group; then there are concepts that
build on this stability and use them to frame newly learned concepts,
which congeal over generations, and are adapted and passed on socially
(we may call this domain knowledge).  This includes specialized
sensors that tokenize, or dimensionally reduce, information intensive
data into compressed conceptual representations\cite{spacetime3}.
Finally, there are concepts, which build on the foregoing, that are
formed over shorter timescales by the observation and introspection of
a single observer (see later figure \ref{fournames}).  The latter is what
we normally think of as learning; however, it is important to remember
that it builds on a stable set of preconditions that have evolved
prior to the current learning episode. We would expect consistent
knowledge to be represented by eigenstates of a memory network.

In approximate order of aggregation, learning may be viewed at these scales:

\begin{itemize}
\item {\em Slow (evolutionary) training} (unsupervised adaptive
  learning): Random variations are whittled away into a set of
  behaviours and memories encoded in a long-term memory, and providing
  the seeds and the framing into which newer ideas are be formed
  through realtime recombination. The hypothesis here is that such
  ideas are naturally associated with the the most invariant aspects
  of space and time, leading to four basic kinds of conceptual
  association, as well as the basic ability to compress extensive
  spacetime datasets into simple tokenized concepts\cite{spacetime3}.

  In a software system, this is represented by hardcoded semantics,
  and specialized sensors, designed to discriminate pre-understood
  concepts, by virtue of their adaptation (e.g. sensors for hot/cold,
  light/dark, movement/no movement, face/no face).

\item {\em Fast (direct) cognitive assessment or context} (unsupervised
  learning and recall): Context changes quickly, and some concepts and
  associations are formed at this timescale.  Each independent
  observer inherits the adaptations learned by previous generations
  and can build on this basis, assembling context from sensory
  experiences, and assessing as emotional state by pattern
  recognition. The sensory channel will eventually be supplemented by
  an introspective channel (see last item). The effect of being
  simultaneously in mind, by observation or thought, leads to
  aggregate clusters, or composite concepts, by co-activation, which
  can be then named as new concepts.

  In software, this represents monitoring and classification of data.
  It might be represented by facial or handwriting recognition
  algorithms for training, etc.

\item {\em Slow (indirect) training} (supervised linguistic learning):
  Given a lexicon of tokens (i.e. a language, however primitive),
  concepts inherited in a compressed form may be passed on from one
  individual to another, e.g. by word of mouth, or as text in a book.
  There is no longer need for direct experience.  A set of concepts
  may thus be remembered in a knowledge bank, such as generational or
  societal memory, and be handed down as domain expertise. This form
  of knowledge acts as a second level of boundary conditions for
  framing and seeding new concepts. The fast cognitive knowledge is
  useless without having these seeds and constraints provide basic
  conceptual anchors.
  In life, the analogy would be the existence of domain knowledge
  that frames our current awareness of a situation.

\item {\em Fast (recurrent) introspection} (emergent adaptation): Once a
  knowledge representation contains a sufficiency of concepts and
  memory representations to be able to form stories without new outside
  stimulus, a system can think about them and associate freely, leading to
  further new concepts. This would be essentially talking to oneself, as it would naturally
share the same linguistic representation as the exterior sharing.

\end{itemize}

At each stage, the key question is how concepts are addressed and
recalled, based on a mixture exterior and interior stimulus. This is
what we mean by the naming of concepts\cite{spacetime1}\footnote{This
  suggests that language would co-evolve with a conceptual
  representation of knowledge, and that utterances, i.e.  statements
  and stories would grow in sophistication along side the knowledge
  evolved in a society of such representations or brains.}.  This is
discussed at length in \cite{spacetime3}.

Finally, no memory system would be complete without processes of
annealing and garbage collection, to even out and clear away clusters
and memories that do not become so important that they dominate over
older ones.

\section{Encoding and retrieving conceptual associations}

Data are singular events, with no a priori inferable interpretation.
The meaning of a single data point can only be promised by its source.
A knowledge system has to be able to integrate the diversity of such
experiences into a stable aggregation, channelling perceived conflicts
into contextualized differences, thus preserving them without
neutralizing them. In a sense, diversity is a prerequisite for
knowledge, and integration leads to a scaled form of hashing.

\subsection{Direct primary cognition}

\begin{figure*}[ht]
\begin{center}
\includegraphics[width=14.5cm]{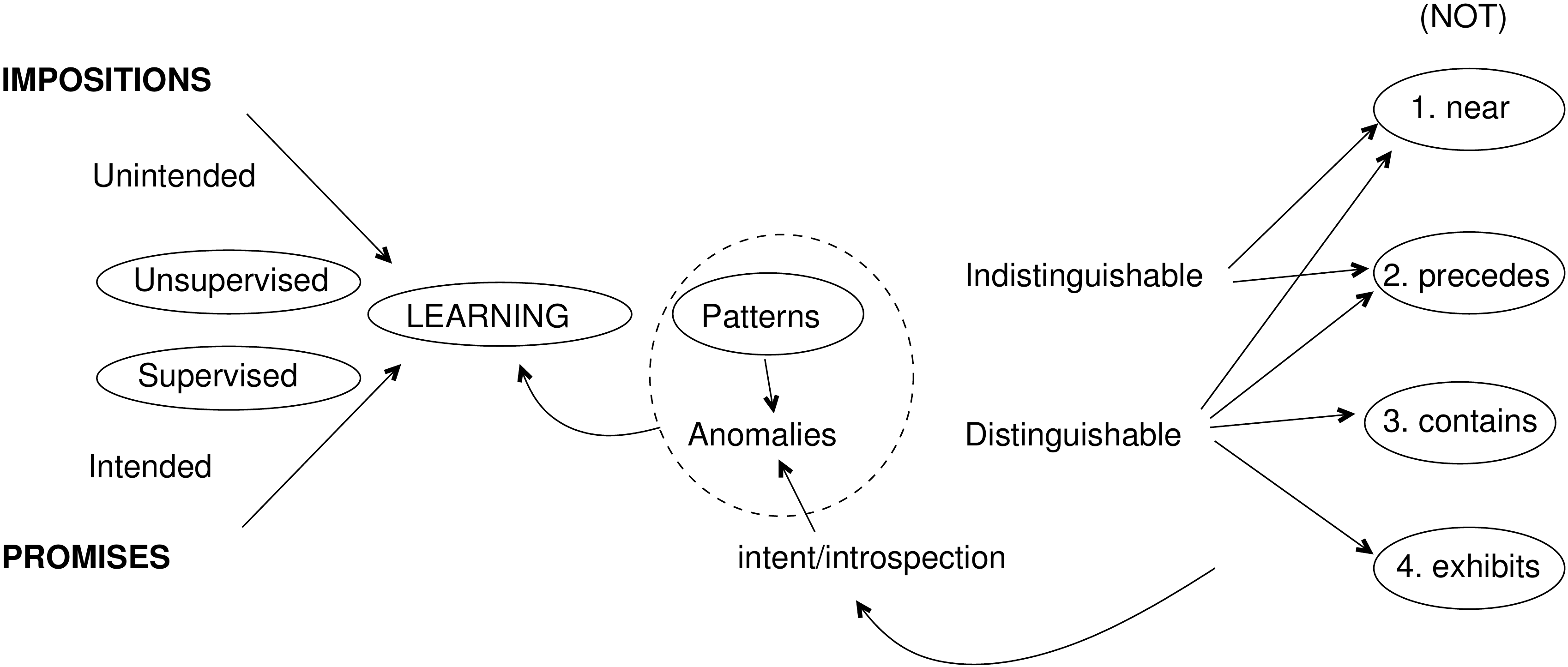}
\caption{\small A concept prism from left to right for association types.\label{prism}}
\end{center}
\end{figure*}

Cognition leads to an association of tokenized concepts through a
process illustrated schematically in figure \ref{prism}. This prism
separates data into a spectrum of four basic associative types, known
as the irreducible types (see table \ref{assoc}).  It mirrors ideas in
linguistics on the classification of nouns.\cite{tratz2010taxonomy}.
The semantics of cognition are very different from the semantics of
experimental observation in science. Ensemble measurement is about
eliminating `self' from experience.  In cognition, observer
subjectivity is a key aspect that cannot be disregarded\footnote{A lot
  of attention has been given over to the algebraic aspects of
  measurement in physics, e.g.  quantum mechanics, but surprisingly
  little attention has been given to the semantics of data, and
  cognitive perception, except in the case of experimental
  error\cite{curd,errors1, errors2}.  The extent to which measurement
  disturbs the system during the act of measurement affects what can
  be promised about a measurement.  Measurables may have the property
  of `compatibility'\cite{schwingerKD}, meaning that measurement of
  one does not influence the measurement of the other.}.

\subsection{Indirect secondary cognition (empathy)}

In machine learning, one tries to fit bulk recorded experience, in all
its detail, directly into a singular neural network, simulating direct
first-hand experience by brute force reconstruction, then adapting its
hardwired nature to match presumed semantics. However, humans are also
able to learn from books and stories passed on verbally: we can learn
from simplified, tokenized representations of knowledge, in which we
do not have to experience every sensation and emotion as the source
did. Even when learning to read books, neural network techniques would
need to read every book into a network in order to train it, rather
than using a short summary of what each book was about. As humans, we
are able to scale knowledge acquisition through summarization, on trust, without
verifying directly from a source. In order to scale machine learning,
machines must also be able to do this.

When training a network by second hand teaching alone (e.g. from books
rather than from practice or experience), we lose realistic sensory or
emotional context, which forms part of our access key to recover the
memories. Thus a model cannot simulate reality from a memory
standpoint.  Some aspects of context can be simulated or described
second-hand, but a realistic emotional state will not normally match
actual first-hand experience.  The ability of a second-hand agent to
{\em empathize} with the concepts and sensations of a primary agent
will therefore play a major role in the ability of pass on
knowledge\footnote{This introspective simulation of emotional context
  may well be the evolutionary purpose of empathy, given the enormous
  value of saving in communication and computational processing.}.  On
the other hand, the compression of such a complex lookup key, into a
more compact address, allows to reorder concepts into new categories,
and to generalize them, overlooking irrelevances based on context.

Cognition therefore ultimately leads to associations between tokenized
(or dimensionally reduced) concept representations, through a
whittling process, described in figure \ref{prism}.  This `prism'
separates data into a spectrum of associative types, which theory
predicts to represent different spacetime characteristics, called the
irreducible types\cite{spacetime3} (see figure \ref{assoc}).  The
conceptual prism bears some similarity with the staged structures of
neural networks; however, neural networks are relatively expensive compression
algorithms, that are not well suited to the latter `expansive' stages
of reasoning, where the bulk of input data would make reasoning too
slow.

\subsection{Tuple form of semantic graph}

Let us assume that the result of a learning process is a set of tuples of the form:
\beq
(C_1,\tau,A_+,C_2,A_-,\chi^*)
\eeq
where $C_1, C_2$ are dimensionally reduced concept tokens, $A_+,A_-$
are forward and backward associations of type $\tau \in \{1,2,3,4\}$ and $-\tau$ respectively,
and $\chi^*$ context labels.  
These tuples represent links or edges in a graph $\Gamma(C,A)$
whose nodes $C$ are conceptual tokens, and whose edges $A$ are associations.
This summarizes an associative
relationship implied by an observation. We can also add a weight or
relative strength for each tuple and a timestamp for when it was last
updated (which in a full knowledge equilibrium, with garbage
collection, would be equivalent).  All of these elements are strings.
For example, highlighting the main features of the tuple, we may represent
a simple qualitative description as a tuple: \small
\begin{center}
\begin{tabular}{|r|l|}
\hline
{\bf doctor} & concept\\
4 & ST type\\
{\bf promises} & forward association\\
{\bf identity credentials}&next-concept\\
is promised by & reverse association\\
{\em patient doctor registration} & context\\
\hline
\end{tabular}
\end{center}
\normalsize In other words, we know that a doctor promises to have
identity credentials, and we are thinking about patient-doctor
registration.  This association is of type 4, on the prism.  We can
apply the same approach to turn numerical data into qualitative
linguistic representations in this format, so \beq X = 47 \eeq might
translate into:
\begin{center}
\begin{tabular}{|r|l|}
\hline
{\bf X}, & concept\\
4 & ST type\\
{\bf has value}, & forward association\\
{\bf identity credentials}&next-concept\\
is the value of by, & reverse association\\
{\em variable assignment} & context\\
\hline
\end{tabular}
\end{center}
In other words, the token $X$ has a value of 37, and we are thinking about variable assignment.

What is noteworthy about the tuples is the absence of data types for
the tokens, translating in an untyped graph.  This is not a
conventional data representation; rather it is a symbolic (purely
linguistic) representation. This must be so, since semantics can only
be extended by reference to other concepts, and these must be rooted
somewhere in language. All concepts are thus symbolic strings,
and their interpretation lies in the way they are associated.

Given such a tuple, nodes and edges are created for $C_1, C_2, \chi^*$ and a
root node $C_{\rm all}$ directed associative edges are added for: \beq
C_1 &\assoc{A_+,\chi^*}& C_2\\
C_2 &\assoc{A_-,-\chi^*}& C_1\\
C_1 &\assoc{ST~ 3}& \chi^*\\
C_2 &\assoc{ST~ 3}& \chi^*\\
\chi^* &\assoc{ST~ -3}& C_1\\
\chi^* &\assoc{ST~ -3}& C_2\\
\chi^* &\assoc{ST~ 3}& C_{\rm all}\\
\eeq With this linkage, we can determine which associations $A_i$ are
relevant to which contexts $\chi_j$ and also which concepts $C_k$ may
be activated by contexts $\chi_j$.  The encoding of concepts in this
way, linked by associations, forms a recursive structure, which is described below. This is not regular in
the manner of a Cayley tree: the result is closer to a `semantic small
worlds' graph \cite{west1,berge1,watts1}.

\section{Naming of concepts and the structure of a name}

Semantic interpretation is essentially about the naming of things, in a representation
meaningful to the observer.  A name is a pattern (behavioural or
symbolic) that represents a concept. Names are most useful if they are
shared between multiple agents, so that concepts can be communicated
and committed to shared (societal) memory, and be used to explain
(not merely recognize) phenomena.  Thus, the way in which we name observed patterns is of
vital importance to their interpretation by other agents. This includes numerical names
such as coordinate tuples, e.g. $(1,2,3,4)$, $(x,y,z)$, etc.

Without a consistent pattern representation for concepts (which we may
define to be a language) there could not be recall of memory, and thus
concepts would be useless and irretrievable. Thus language is a
necessary condition for semantic interpretation. In neural network
approaches, the dimensional reduction of a number of inputs to a small
number of outputs is only meaningful when one can clearly and
unambiguously identify the output channels with named concepts.  This
too is a simple language transformation\footnote{Note that this
  suggests that language is not a rule-determined system, but a
  pattern-based one, i.e. that any rules we identify are post-hoc
  rather than generative.}.

\subsection{Associations and their aliases in context}

As argued in \cite{spacetime3}, the fundamental basis for conceptual
association has its origins in the structure of spacetime, represented
by four types of relationship $\tau$ (being close in space and time, being in a
bounded area, etc). We then give specific associative meaning $A(\tau)$ to these different
spacetime coincidences by naming associations according to abstracted
concepts. Concepts and associations are nodes and edges of a graph:
\beq 
C_1 \assoc{\cal A(\tau)|\chi} C_2 
\eeq 
where the edge of the graph depends on a fundamental spacetime type,
whose qualitative description is given by one of the four irreducible
associations documented in table \ref{assoc}, and their negatives.

The conditional parameter $\chi$ represents a context set of terms,
under which this edge is active. Since context changes, by definition,
as fast as we can think, or the environment can be perceived, the
elementary tokens associated with context must be expected to be as
large in number as the number of invariant sensory perceptions of the
observer, and the most rapidly changing symbols in the knowledge
representation. For this reason, we could expect primary context
characterizations to be hardwired, i.e.  built into the functional
nature of the agent, by specialized eyes and ears, face cells and
place cells, and so on\cite{gridcells}.  Secondary characteristics could be learned over
time, provided they remain simple enough to be activated quickly. It
is unlikely that context would be expressed by complex concepts.
However, it could be that this is the role of emotions: sensations as
complex as sensory perception, but without strong rational
linkage\cite{lisa98283}.
\begin{table*}[ht]
\begin{center}
\begin{tabular}{|c|c|c|c|}
\hline
\sc ST Type & \sc Forward & \sc Reciprocal & \sc Spacetime structure\\
\hline
\hline
&is close to & is close to & contiguity\\
&approximates  & is equivalent to &\sc PROXIMITY\\
ST 1&is connected to & is connected to&``near''\\
&is adjacent to & is adjacent to&   Semantic symmetrization \\
&is correlated with & is correlated with& similarity\\
\hline
&\sc Forward & \sc Reciprocal & \sc Spacetime structure\\
\hline
&depends on & enables& ordering\\
ST 2&is caused by &causes & \sc GRADIENT/DIRECTION \\
&follows & precedes& ``follows''\\

\hline
&\sc Forward & \sc Reciprocal & \sc Spacetime structure\\

\hline
&contains & is a part of / occupies& boundary perimeter\\
ST 3&surrounds  & inside & \sc AGGREGATE / MEMBERSHIP \\
&generalizes & is an aspect of / exemplifies& ``contains'' / coarse graining\\
\hline

\hline
&\sc Forward & \sc Reciprocal & \sc Spacetime structure\\
\hline
&has name or value &  is the value of property& qualitative attribute\\
ST 4&characterizes &  is a property of &\sc DISTINGUISHABILITY\\
&represents/expresses &  is represented/expressed by & ``expresses''\\
&promises && Asymmetrizer\\
\hline
\end{tabular}
\end{center}
\caption{\small Examples of the four irreducible association types, characterized by
their spacetime origins, from \cite{spacetime3}. In a graph representation, `has attribute' and `contains'
are clearly not independent, so implementation details can still compress the number of types.\label{assoc}}
\end{table*}

Multiple aliases, or alternative interpretations, of the four spatial
relationships are possible, indeed they are encouraged in effective
communication for expressivity and qualification. For example, type 1
(proximity) may be interpreted as adjacency, approximately equal to,
close to, next to, etc. Type 2 (linear sequential order) may represent
time, or unidirectional ordering, causation, dependency, etc. Type 3
(containment) may represent membership in a group, generalization of a
collection of concepts, location inside or outside a perimeter, etc.
However, we should also be cautious that informal association of
linguistic metaphor also leads to confusions about the appropriate
classification of meaning under the irreducible types, as
interpretation by metaphor is fluid in human language\cite{unfolding}.

Associations must retain their specific interpretation on every link,
but also the generic type, which affects the way we parse the
structure and hence reason about the associations. Meaningful
narrative is usually driven mainly by type 2 (causally ordered
sequences of happenings or reasoned steps), embellished by type 4
discriminating properties and lateral reasoning of types 1 and 3.

\subsection{Scaling of semantics}

Semantically enhanced concepts may be derived from the aggregation of
primitive concepts, by associative clustering.  Each cluster may act
as a new compound agent, through its representative hub (whose name
represents the aggregation), by making new associative links.  The hub
acts as a gateway to the qualified concept's associative network, much
as a router is a gateway between a subnet and the larger Internet.
Each concept is therefore an agent, at some scale of aggregation, with
a unique name representing its meaning.  In addition, the names of
contextualized associations, within a cluster, and without, add the
specificity of concepts expressed.

Based on trial and error experience, it seems that the most
comprehensible or `natural' linguistic naming strategy is one that
follows the scaling of agency, i.e. aggregation, as described in
\cite{spacetime2}.  With a scaled approach, one combines the names of
interior promises to yield a compound name for a collective concept,
and exterior promises then express the associations between the
composite concepts\cite{spacetime2}. In linguistics this is called
{\em compounding}.  The scaling is compatible with patterns observed
in cognitive linguistics \cite{langacker1}, indicating that language,
as we understand it, forms a good network structure for concept
representation\footnote{It should probably be considered a tautology
  that linguistic structure and conceptual networking are
  complementary.}.

From a cognitive perspective, the way we turn data into concepts depends on
the spacetime boundary conditions imposed by exterior environment.
The semantics of data are different depending on whether samples
originate from from a single source or from a collection (statistical
ensemble) of sources.  In information culture, people sometimes talk about {\em pets versus
cattle}, meaning unique instances with names and labels, versus herds
without distinct identities. In data terms, these are singletons and
ensembles.

\subsection{The role of approximation in scaling invariance}

In a batch experiment, where multiple samples contribute to a
collective impression, all data points within an ensemble are
considered to be equivalent, and with homogeneous semantics. In other
words, we choose (as a matter of policy) to overlook any differences,
and map them all to the same name.  If a dataset is a stream of scalar
values $q_i$, then every $q_i$ would be considered interchangeable.
The indices $i$ may be relabelled and the results will be preserved.
The properties that are derived from the dataset must therefore be
invariant under relabellings of the array index $i$. This is how {\em
  invariant representations} are formed. Such invariant
representations are crucial to avoid an explosion of
contextualization, which would be expensive and would render concepts
non-reusable. In order to learn from the past, a cognitive agent
must transform ephemeral samples into invariant concepts efficiently.

\begin{figure}[ht]
\begin{center}
\includegraphics[width=6.5cm]{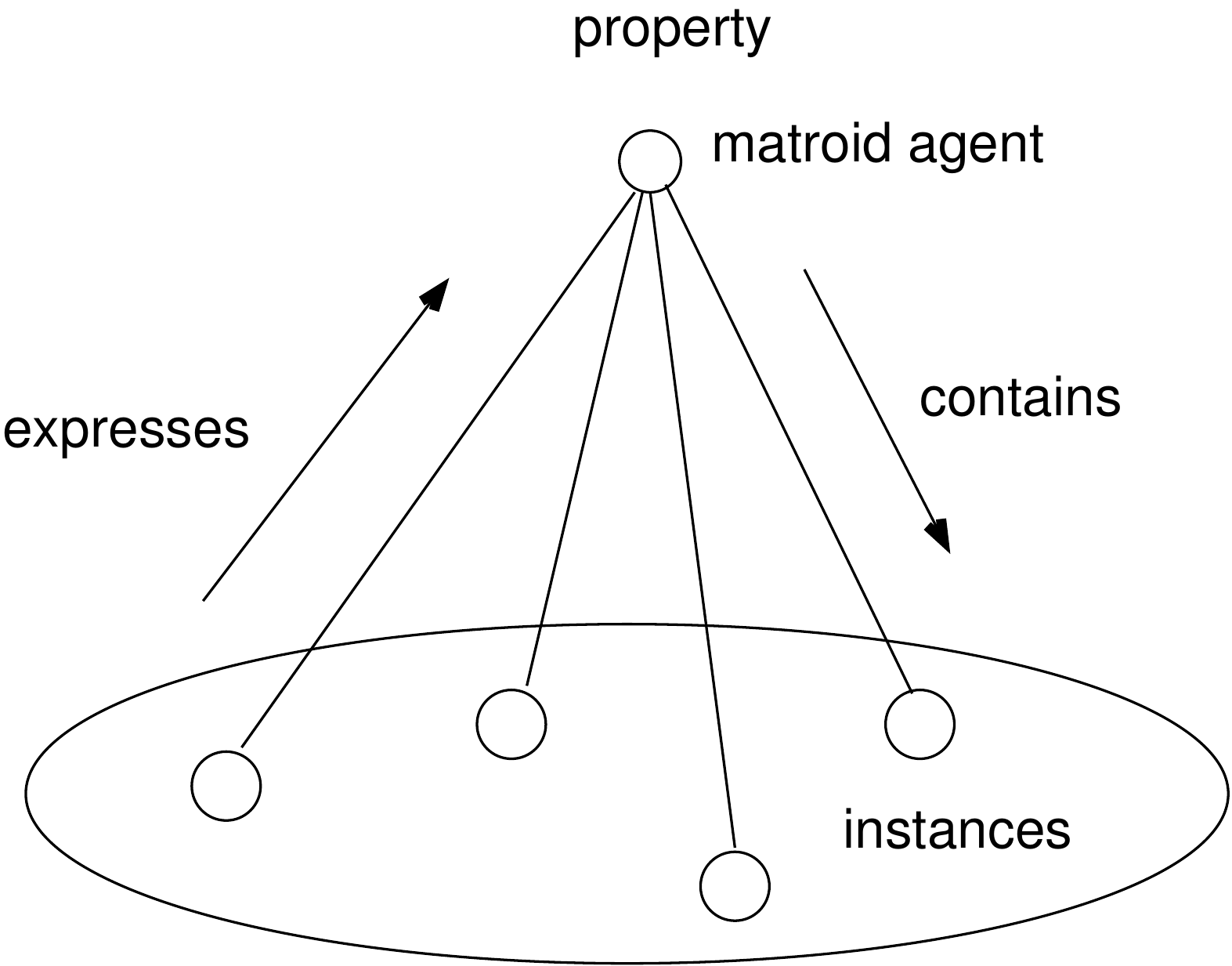}
\caption{\small The semantics of representing aggregation. If we group aggregated
  properties by a linking them centrally to a singular basis
  agent\cite{spacetime1,spacetime3}, then attribute expression is
  effectively the inverse of set membership, and the basis agent alone
  represents the collective ensemble. Ad hoc clusters not have this
  property.\label{central}}
\end{center}
\end{figure}

\begin{figure}[ht]
\begin{center}
\includegraphics[width=6.5cm]{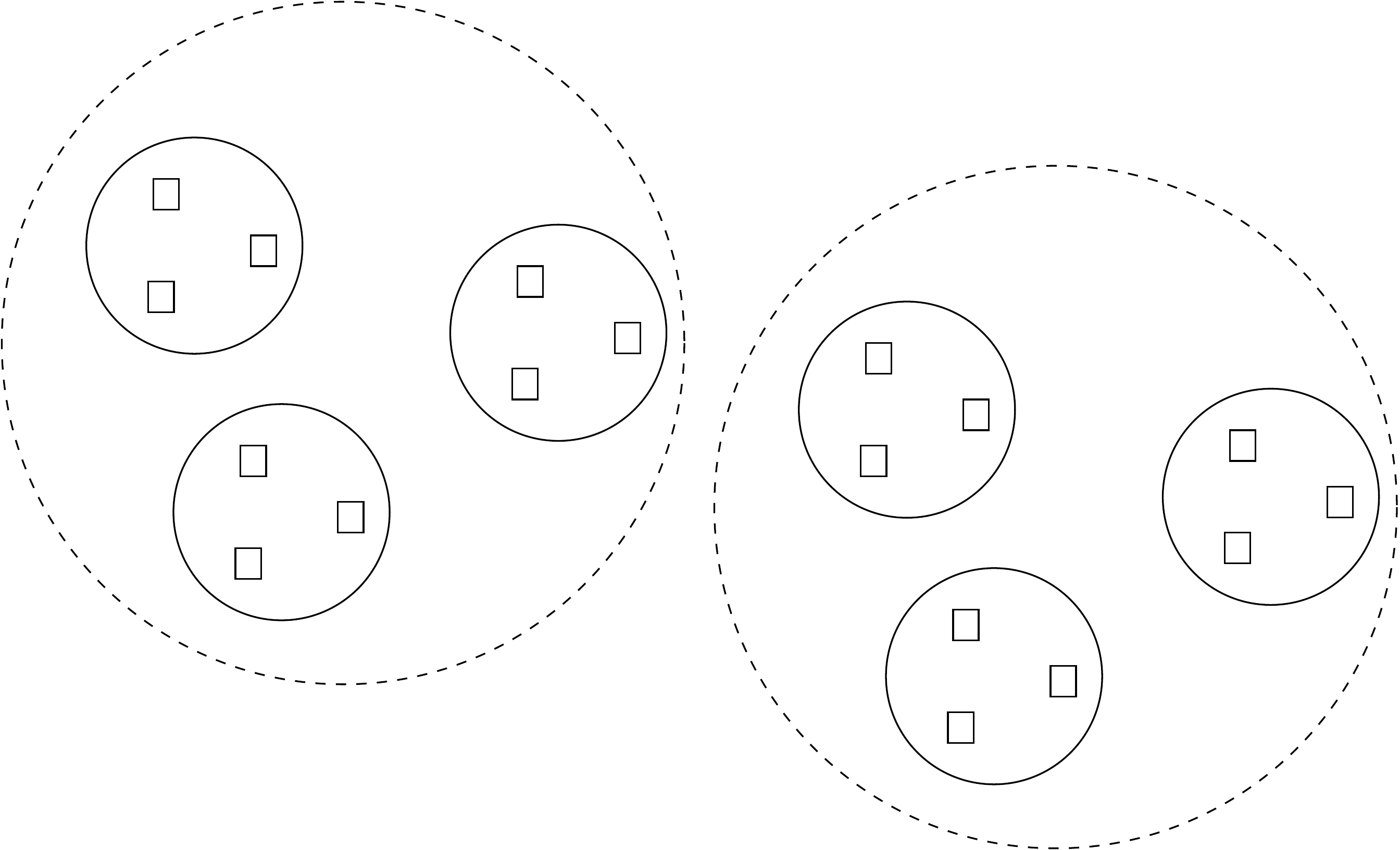}
\caption{\small Indistinguishable agents at different scales.\label{multiscale}}
\end{center}
\end{figure}

For each of the irreducible spacetime association types, there is an
interpretation of the meaning of aggregation, which promotes invariance:
\begin{enumerate}
\item {\em Proximity clustering or approximation} (spacelike or semantic convergence):
When named concepts are close together either in the real external world, or in the interior representation,
then they form a labelled proximity cluster. The nature of the proximity may be physical, literal, or semantic
depending on the label. The encoding is the same in all cases, up to a label. See figure \ref{cluster4}.
\begin{figure}[ht]
\begin{center}
\includegraphics[width=5.0cm]{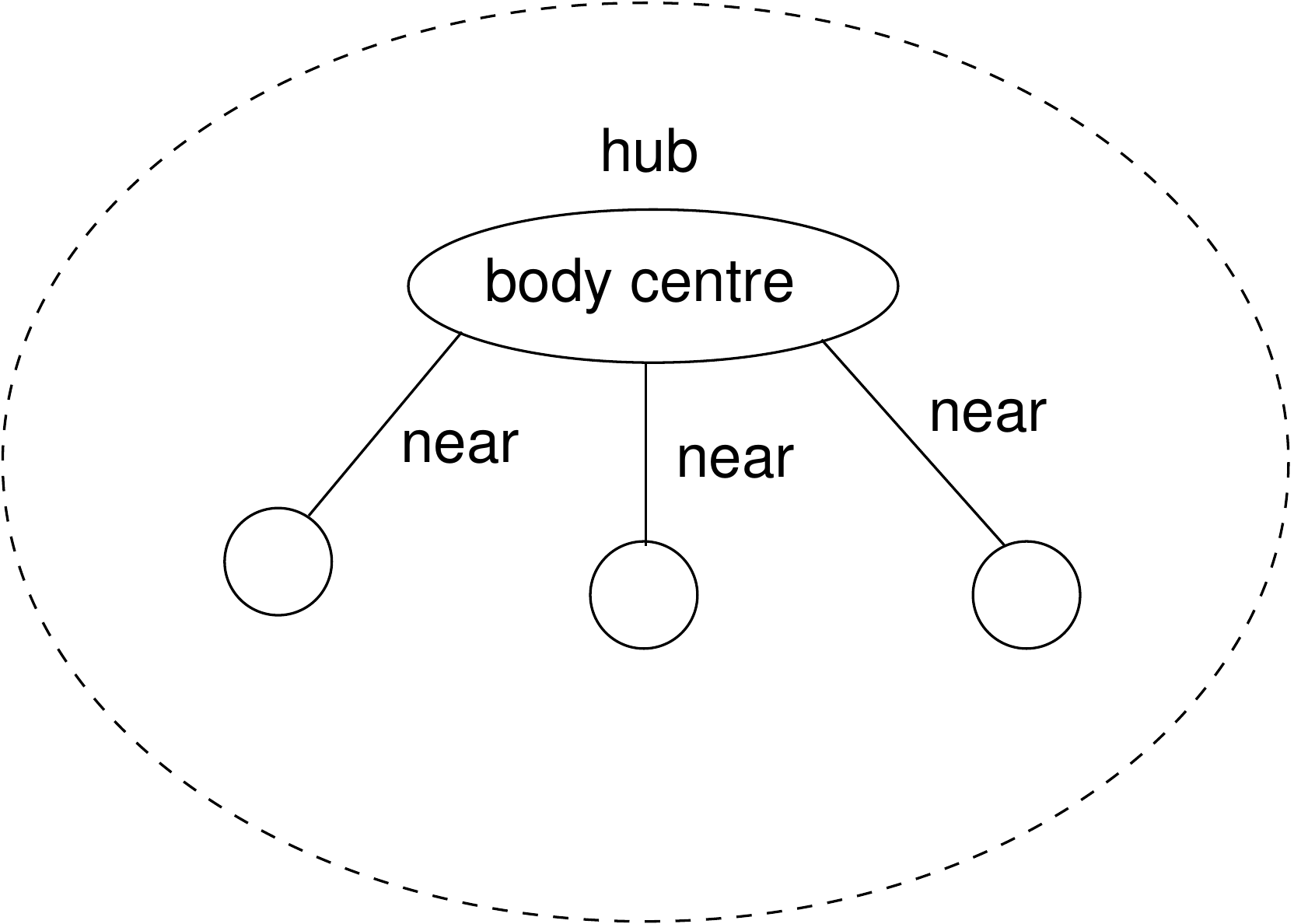}
\caption{\small ST Type 1 Proximity cluster is accretion.\label{cluster4}}
\end{center}
\end{figure}

\item {\em Causal aggregation} (timelike convergent dynamics):
When dependencies come together to a head, so that a single outcome may be a causal determinant of multiple
sources. This is a definition of the arrow of time. See figure \ref{cluster2}.
\begin{figure}[ht]
\begin{center}
\includegraphics[width=5.0cm]{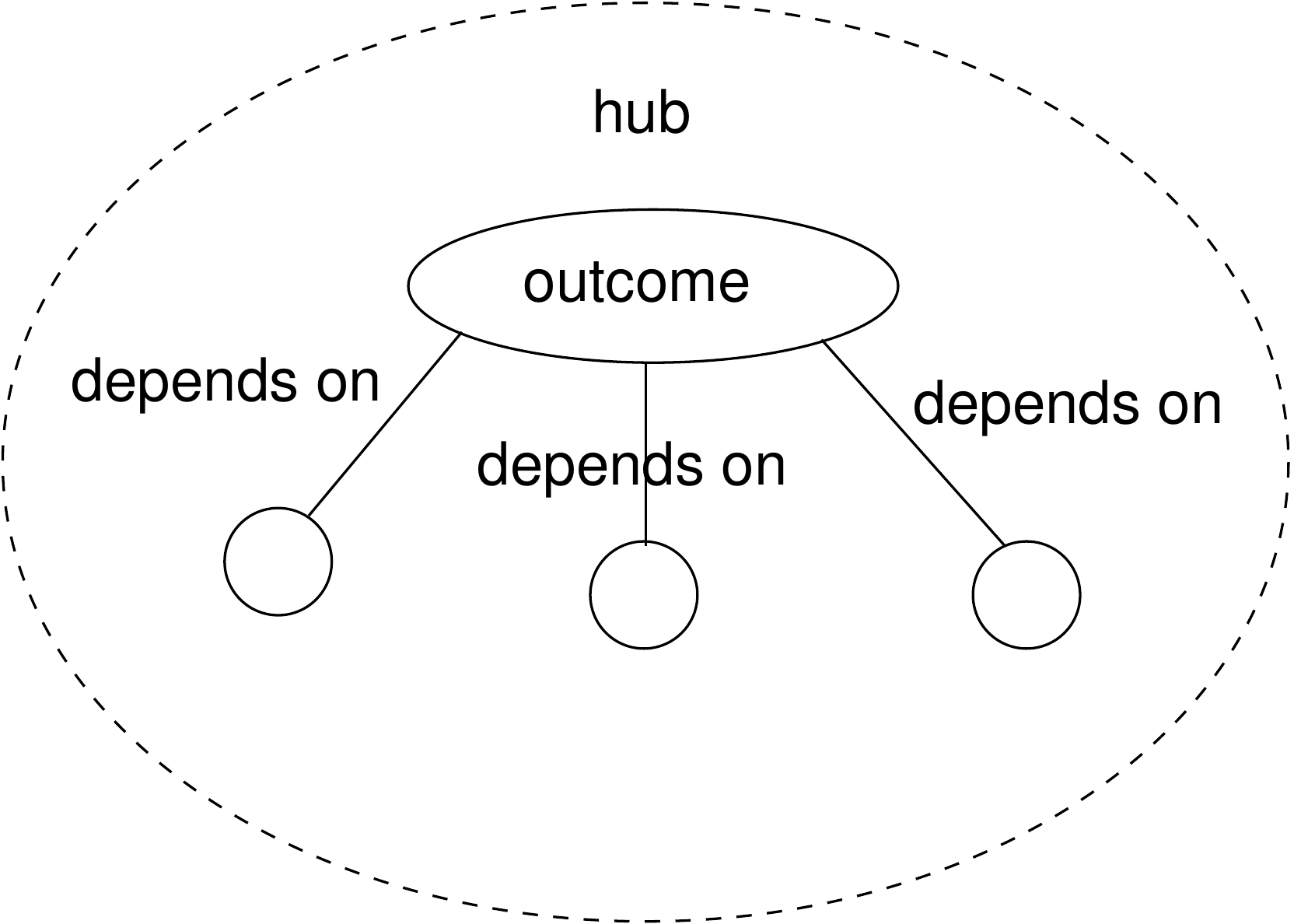}
\caption{\small ST Type (2) Causal cluster indication causation, like in a fault-dependency tree.\label{cluster2}}
\end{center}
\end{figure}

\item {\em Symmetry aggregation (membership or spacelike homogeneity)}:
When concepts are joined to a single hub that represents their collective identity,
by a `contains/contained by' relationship. See figure \ref{cluster1}.
\begin{figure}[ht]
\begin{center}
\includegraphics[width=5.0cm]{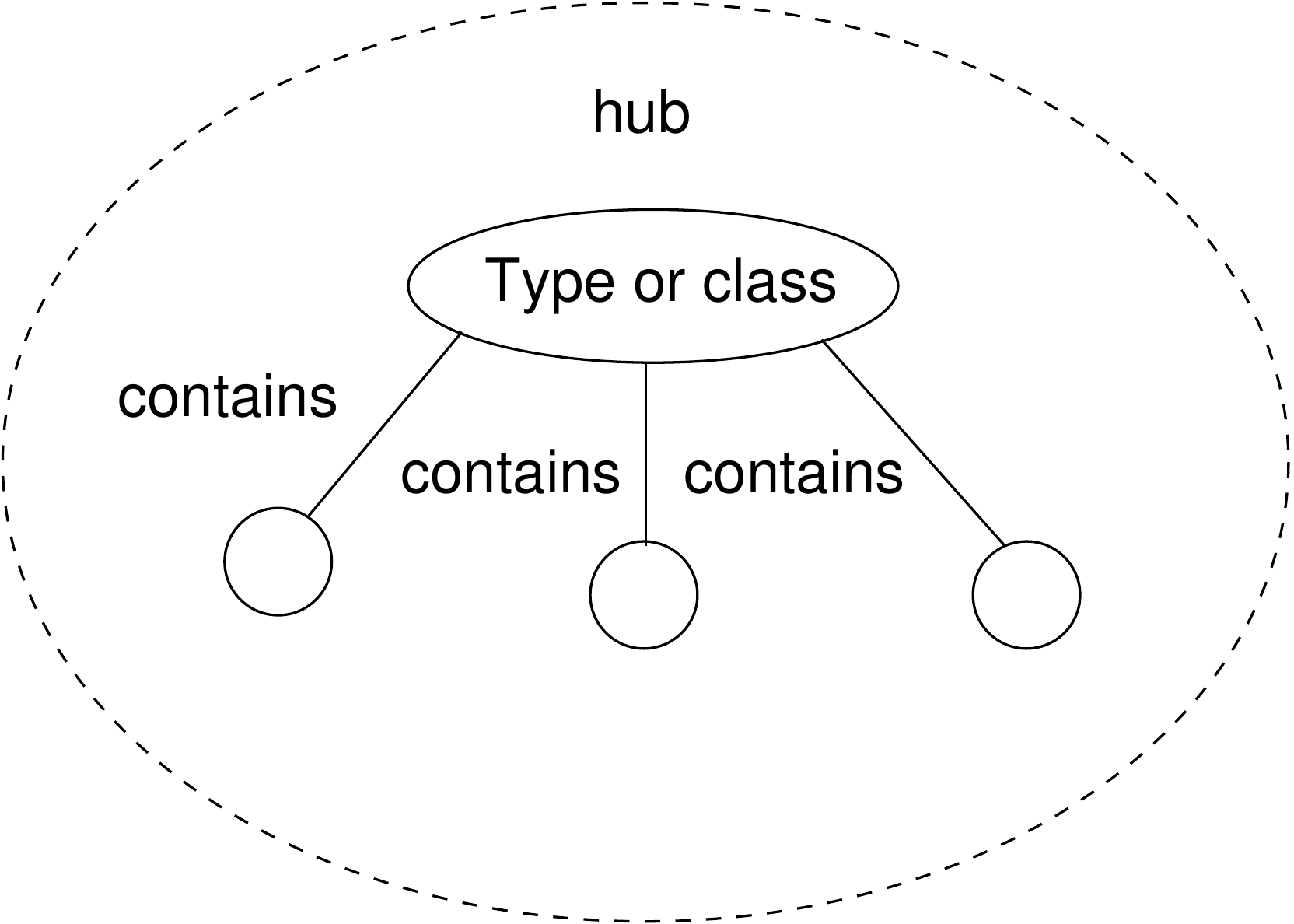}
\caption{\small ST Type (1) Symmetry cluster or membership class\label{cluster1}}
\end{center}
\end{figure}

\item {\em Qualification by attribution} (semantic qualification):
When the name of a concept is qualified by a context (sometimes called a namespace), and possibly a role (like verb, noun etc),
For example, the multiple interpretations of a word like ``tar''
\begin{itemize}
\item Tar as a noun in the context of waterproofing
\item Tar as a noun in the context of computing
\item Tar as a verb in the context of painting
\item Tar as a verb in the context of characterization
\end{itemize}
must be distinguished using the hub construction for attribute association. 
This construction is a generic (pattern) form of what we call a schema or datatype in computing.
It is recognizable without any specially arranged protocol, just from the observed associations.
See figure \ref{cluster3}.
\begin{figure}[ht]
\begin{center}
\includegraphics[width=5.0cm]{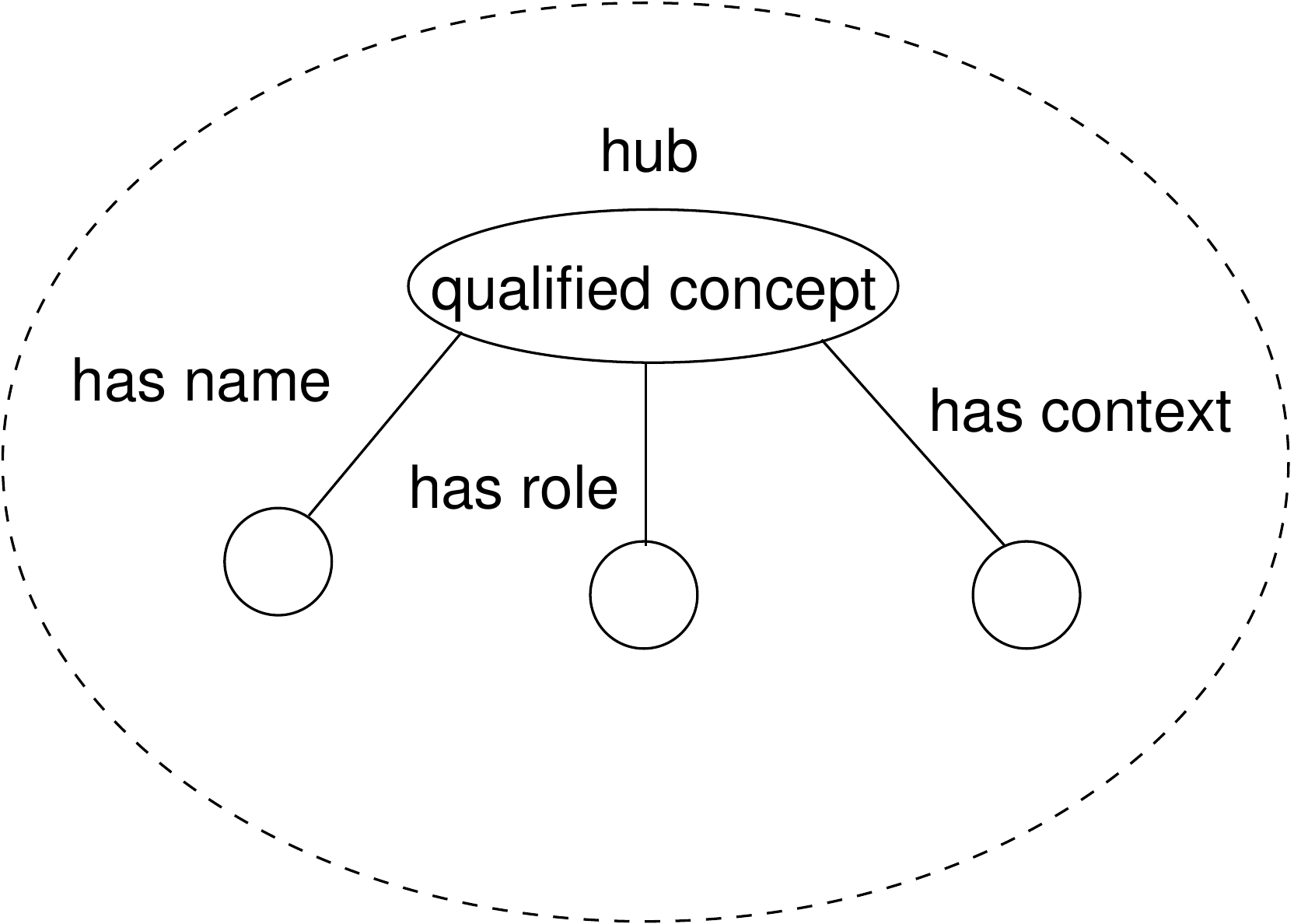}
\caption{\small ST Type (3) Semantic qualification clusters describe the kind of concept and its context.\label{cluster3}}
\end{center}
\end{figure}
\end{enumerate}
The scaling of conceptual agency, in these representations, is a direct application
of the promise theory\cite{spacetime2}, summarized here according to
the four irreducible spacetime attributes elucidated in
\cite{spacetime3}.
The parsing algorithm can now be be remarkably simple because the
associations are organized around the most elementary spacetime concepts,
or distinction in location, order, or expressed properties.

\section{The matroid naming construction}\label{compounds}

In earlier work\cite{spacetime1}, I showed how a promise view of
structure motivated an identification of concepts as spanning sets, or
matroids over a labelled graph. The most convenient representation of
these is to bind all related clusters to a hub node, whose name is the
collective name of the cluster\footnote{This is broadly analogous to
  the way we name a compound data schema object in computing. The hub
  has the name of the object in its entirety, and the members have the
  names of the associated items. The association types are analogous
  to the data types. Where the analogy breaks down is that matroid
  hubs can name arbitrary collections of data, they do not form
  mutually exclusive, non-overlapping memory segments.}.  This offers
a scalable approach to spanning concepts\cite{spacetime1}, with
upwardly extensible uniqueness.  By associating a collection of
associated concepts with a single point of address, we throw a kind of
perimeter around the spokes and rim emanating from the hub to form a
single new concept. Such clusters may overlap without limit, thus this
is not a mutually exclusive branching
process\cite{burgesskm}\footnote{Concepts may be aggregated without
  limit for qualification, but they cannot be subdivided without
  limit. Thus taxonomies and ontologies as branching processes are not
  scalable to adaptation.}.

Unlike a hashing approach to unique naming, uniqueness can be assured
by combinatoric compounding of names.  The collection of named
concepts in a cluster is accessed through a {\em single node hub}
which acts as a gateway for the aggregation of parts.  The name of such
a matroid hub (or compound name) represents the name of the composite concept,
formed by the internal association of interior concepts.

\subsection{Nominal compounds and orphans}

\begin{figure}[ht]
\begin{center}
\includegraphics[width=6.5cm]{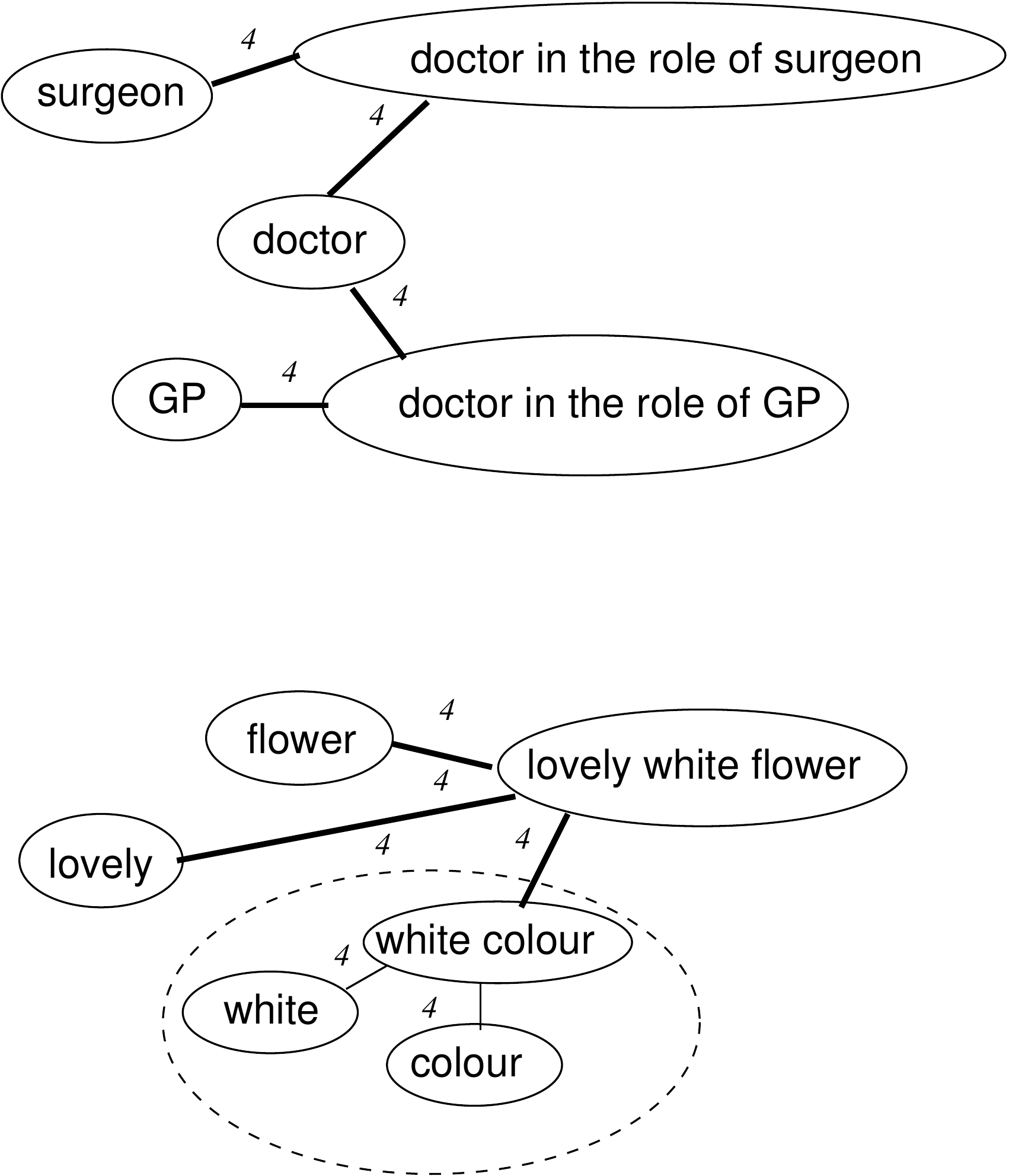}
\caption{\small Under aggregation by a hub, the irreducible types
  effectively collapse into a single type, bound by membership to the
  hub. This kind of central binding is the most invariant construct of
  all.\label{compounds1}}
\end{center}
\end{figure}

Concepts, whether primitive or derived, may therefore be thought of as
`lexical fragments', `syntax fragments', or generally language
fragment, formed by recursive scaling.  A compound is, in essence, a
long name, which contains evidence of the concepts it depends on, and
which is linked recursively to the atomic concepts (see figure
\ref{compounds1}). It is also possible that the compound name forgets
its association to the concepts from which it was derived, and
survives independently through the frequency and associative density
of its use.

\begin{figure}[ht]
\begin{center}
\includegraphics[width=6.5cm]{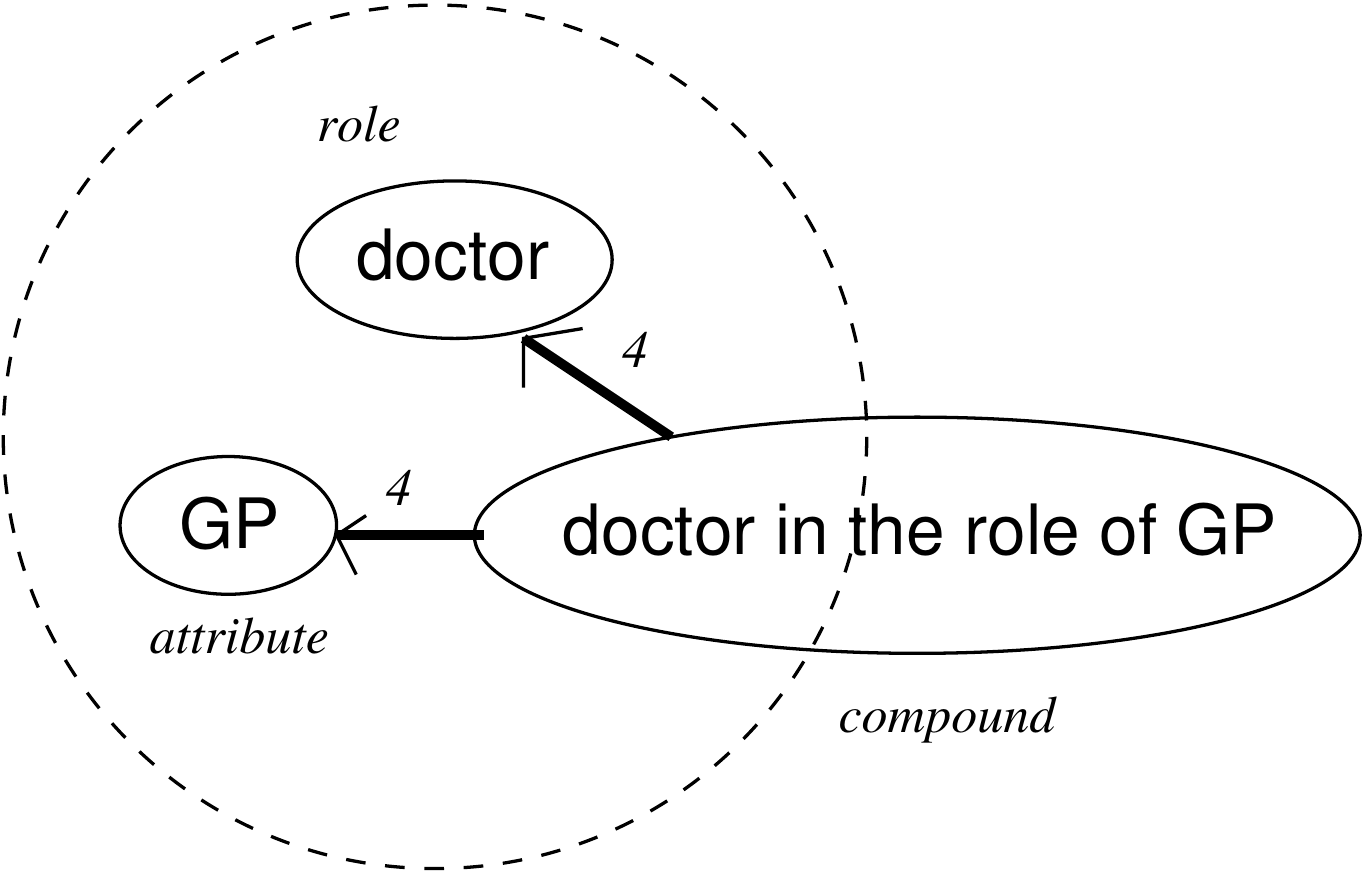}
\caption{\small Compound formation can be reduced to a functional pattern: a representative
name linked to partial concepts (type -1 association) one of which is a main subject header
(like doctor) This is a dimensional reduction strategy, because we can now refer only
to the compound name and drag along all the conceptual baggage of its associations
accordingly. Note, however, that the interpretation of a privileged role for
is sometimes ambiguous.\label{compounds2}}
\end{center}
\end{figure}

The study of compound terminology is a major topic in linguistics,
which addresses the conceptual origins of names we use in language.
For example, the leg of a table can be referred to by a single
compound word or phrase `tableleg', in which the compound has the role
of `leg' (by analogy to an animal leg) and a contextual attribute
`table' (see figures \ref{compounds1} and \ref{compounds2}).  
\beq 
\left(
\text{leg} \revpromise{\rm role} (\text{table-leg}) \promise{\rm qualifier} \text{table}\right) 
\eeq 
This example is straightforward, and fairly unambiguous. The
name reflects the components directly with a conventional ordering,
leading to a simple rule to construct a compound identifier. However,
many compound concepts do not directly reflect their origins, or in
fact forget their origins over time, e.g window (from Norwegian `vind'
and `auge', meaning eye for the wind, clearly before the invention of
glass). A window is neither a wind, nor an eye, yet the word serves
its purpose with a whiff of the metaphysical.  In other cases, there
is no linguistic term for a new qualification, yet we need to
distinguish ambiguous concepts with qualifying context, e.g. consider
`doctor'. A doctor may refer to a medical doctor, a general
practitioner (GP), a surgeon, or someone with a PhD, etc. We could
simply find unique names for all of these, but this approach does not
easily scale; hence, it is normal to describe `namespaces' as
contextual constraints. This can be done by forming a compound name as
a phrase `doctor in the role of GP' for instance (see figure \ref{compounds2}).

It is now somewhat ambiguous what is the role and what is the
qualifier in this construction. It could be argued that `GP' is the
role and doctor is redundant, or that `doctor' is the role and `GP' is
the qualifier.  The distinction is somewhat arbitrary from a
computational viewpoint (this is the danger of `schema' thinking), but it is sometimes helpful to be able to
distinguish a privileged component that describes the behaviour or
function of the term. Through the various experiments conducted during
the course of this work, from topic maps\cite{topicmaps} to the
present model, it seems that the appropriate place for agent role in a
network lies in the associations between pairs of concepts, rather
than in the concepts themselves. If one tries to `type' concepts (in a
typology, taxonomy, ontology, meronomy (partonomoy), etc) one prevents
the free association of ideas by analogy that is so central to human
reasoning. Thus the classic idea of a `data type' for concepts in a
semantic network appears to be simply wrong because it leads to an
over-constrained network with too much uniqueness. The resulting network is too sparse
to allow percolation and hence propagation of storyline\cite{burgessC12,spacetime3}.

Finally, there is the central problem: how do we think up names to
concept clusters?  The naming of clusters seems analogous to the
matter of forming nominal compounds (see figure \ref{compounds1}).
One approach is to try to abandon linguistic meaning and use some form
of unique hash. This approach is used in natural language
processing\cite{hashtopics} with some claimed success. However,
hashing is a one-way transformation, which is unsuitable for our
associative map. To create an interactive system of knowledge, we need
to be able to read back concepts in a form parsable by humans.

The matroid hubs, with their compound names, are binding points, or
confluences of association, which act as semantic correlators, i.e.
authoritative anchor points, expressing uniqueness.  If we follow
through this approach consistently, then conceptual specificity
increases upwards with the level of aggregation, rather than downwards
with the depth of branching, as in a taxonomic hierarchies (i.e.
branching processes).  This leads to a bottom-up approach to
conceptualization, which is entirely compatible with the modern
genetic understanding of phylogeny and phenotype expressed through the
aggregation of genetic flavours.  This is also motivated by a
promise-oriented interpretation. Mirroring the proposed usage approach
of language formation in cognitive linguistics\cite{tomasello1}. It is
a chemistry of semantic atoms, mixed into distinct molecular
combinations, with differentiated functional meaning.

\subsection{Concepts, associations and the scaling hierarchy}

The end result of a cognitive process is to tokenize complex data
inputs into tokens (concepts), which summarize the data in an {\em
  invariant representation}; this might be approximate, but it is
proportionately immune to variations, and to link these together by
association. In promise theory language, concept tokens are agents
(nodes in a graph), and associations (graph edges) are promised by
these agents.

There is a ladder from sensing to conceptualization, which proceeds by
aggregation (over time and space). This involves work, and thus it
takes increasing time to process complex associative concepts.
\begin{figure*}[ht]
\begin{center}
\includegraphics[width=13.5cm]{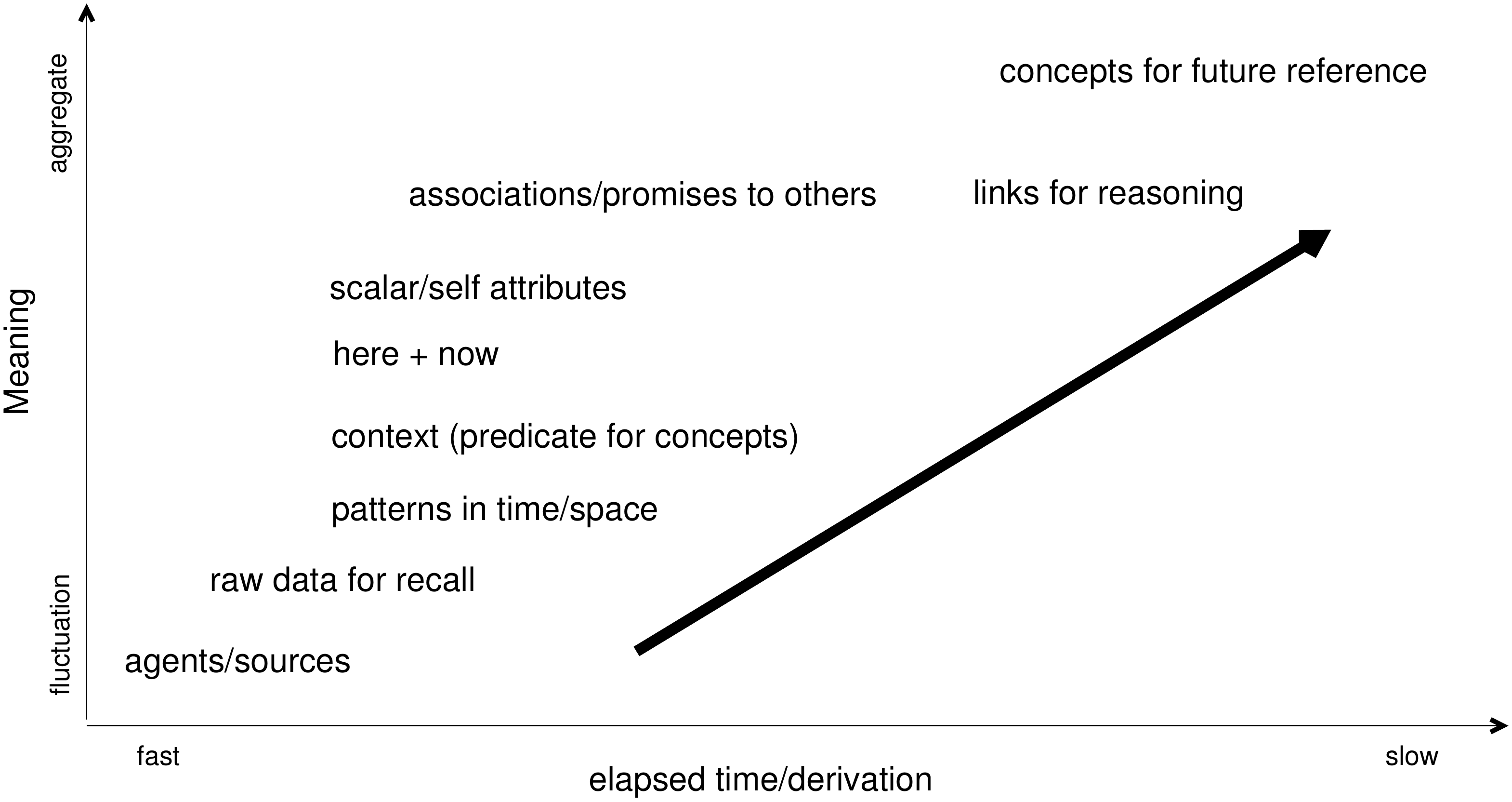}
\caption{\small Climbing the knowledge ladder takes longer and longer
  to advance towards sophisticated concepts, from fast sensing to pondering
  concepts, because of iteration and aggregation (learning).\label{meaning}}
\end{center}
\end{figure*}
Concepts are clusters of agents.
The basis of interpretational semantics lies in the following observations:

\begin{law}[Coincidence and concurrence]
  That which occurs together (concurrently, coincidentally, i.e.
  locally, or in the same context) is associated. Thus context belongs
  to associations, not to concepts. Measurements at different locations
  and at different times are not necessarily causally related, but may
  be correlated.  The naming of the association is undetermined.
\end{law}

\begin{law}[Semantic stability]
Averages stabilize semantics by decoupling from temporal and spatial approximate regions.
  Data sampled and combined from different sources (locations) should
  be treated as ensemble averages, in a single concurrent experiment.
  The meaning of time and location are lost in these averages, and
  semantics are stabilized by this decoupling.
\end{law}

\begin{law}[Scaled agency]
  Semantics (interpretations) arise from the naming of associated
  forms and patterns, in space and time, and across multiple scales.
  Names assigned become invariant concepts.
\end{law}

\begin{law}[Constant semantics during observation]
  Measurements are comparable if they have the same semantics.  In
  cognition, observations are not comparable in an experimental sense,
  unless we deliberately overlook certain information. Thus semantic
stability depends on what information we throw away.
\end{law}
The last two are circularly self consistent.

\subsection{The learned context channel}

During activation, inputs might activate an elementary context token,
and in turn, several such atoms might activate an association,
somewhat in the manner of a crude neural network (see figure
\ref{prism}).  The relative activation level of an association is thus
the integral of the weights of its incoming edges. A triggering policy
could also be employed for generality and tuning; however, we can only
simulate this spatial interference process on von Neumann computers..

Context is used to frame the circumstances under which associations
and concepts are identified. Striking the balance between a context
that strives for a coordinatized precision and one that is maximally
invariant (for repeated `context-free' use) is the central challenge
of building an {\em addressing system} for semantics and knowledge\cite{spacetime3}.  In the
implementation of Cellibrium, it is thus assumed that:
\begin{itemize}
\item Context is attached to associations, not concepts. A
  co-activation of an associative bond needs only be a set that
  activates both ends, but the specific members may differ at each end
  by evolution of contextual awareness of the cognitive agent.

\item Limiting the amount of nominal information in context is a priority, else the memory
requirements for knowledge would be divergent.

\item All associations belong to one of the four spacetime association types\cite{spacetime3}:
\begin{enumerate}
\item An approximate similar location in space.
\item A common cause or outcome, convergence of path in past or future.
\item A common enclosure or membership.
\item A shared property, leading to implicit membership (e.g. all kinds of doctor in figure \ref{compounds1}).
\end{enumerate}
In aggregation by symmetry, all these collapse to type 3 in practice, i.e. grouping
of indistinguishable members\cite{spacetime3}.
\item Context describes something analogous to recent browsing history:
\begin{enumerate}
\item The intent that led to the association being made (interior state).
\item The emotional and rational state of the agent's recent thinking (interior).
\item The unintended circumstances in which the agent finds itself when making
the association (exterior state).
\end{enumerate}
In other words, anything we happen to be thinking about or involved in can be a context for knowledge relevance
(see figure \ref{intextfig}).

\end{itemize}

\subsection{Interior and exterior associations of concept clusters}

\begin{figure}[ht]
\begin{center}
\includegraphics[width=8cm]{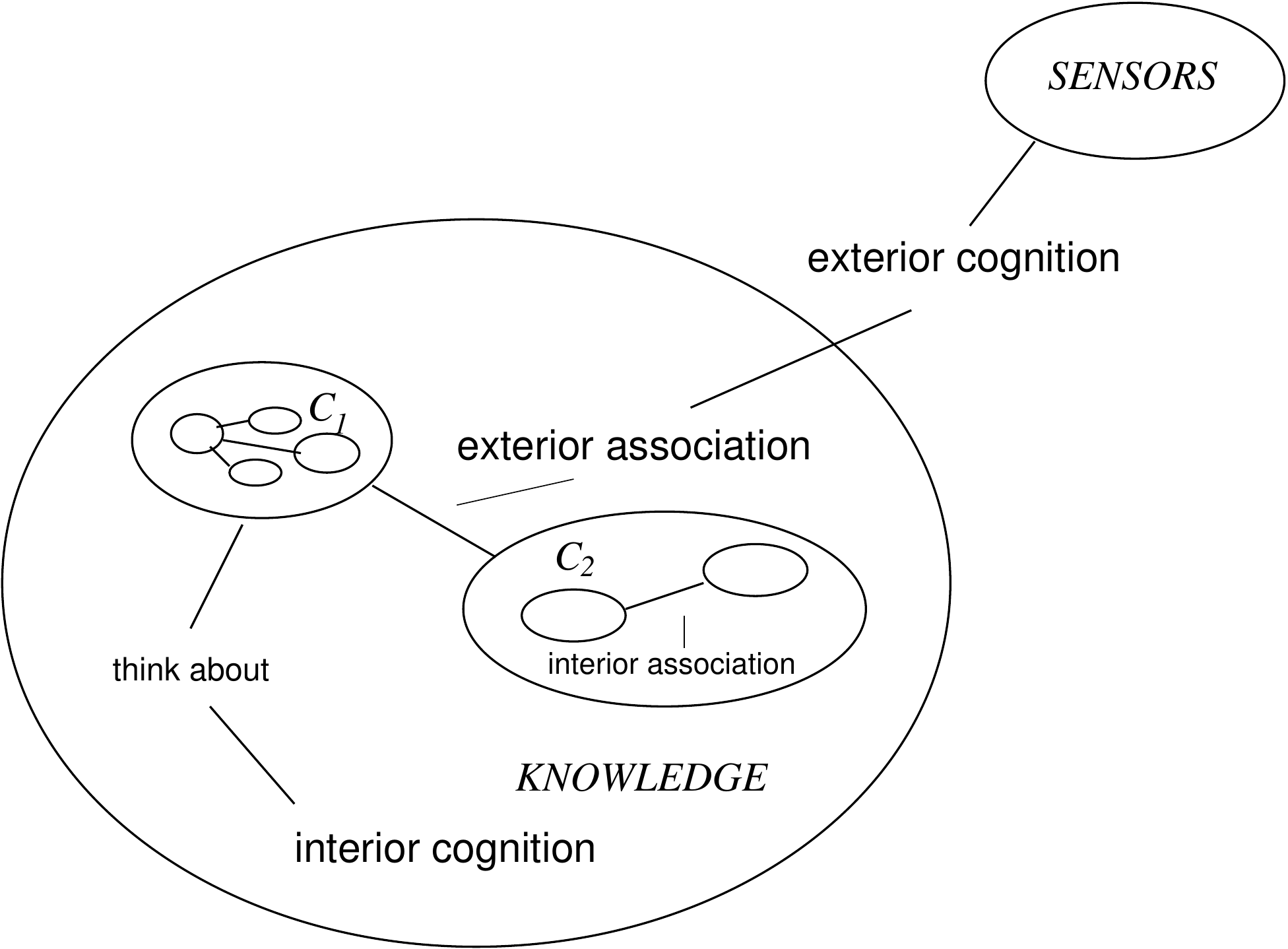}
\caption{\small The scaling of association and cognition. Context labelling of associations may be interior to compound concepts, or exterior between concepts. Context acquisition, by cognition, may be interior to an agent observer (intentional or introspective) or exterior (sensory or unintentional).\label{intextfig}}
\end{center}
\end{figure}

What is expressed externally by sensors plays a different role than what is
expressed internally by introspection, as we saw in section \ref{compounds}.  Interior
and exterior properties continue to play a role in distinguishing.
Figure \ref{intextfig} illustrates this. Both interior and exterior
contexts may play a role in addressing, i.e. labelling and retrieving,
concepts (see section \ref{retrievecontext}).

There remains the question how we should use context to activate or
deactivate story pathways. In an artificial system, especially one made for a
domain specific purpose, the information content, or semantic
complexity, inherent in context annotation during training might be
too small to give a reasonable chance of discrimination of relevance
in real-world circumstances.  Unless there is a sufficient
density of discriminators, we will either fail to find possible pathways
(lateral thinking) or we will include too many, losing the point of
context as a refinement criterion.

We might rank pathways by an activation score for how each node in a
story overlaps with the current context.  Alternatively, we might rank
an entire story along its path. The latter would bias in favour of long
stories\footnote{If we could find a normalized `probability' to sum
per unit story length, then we could build a score analogous to the
Shannon entropy, in which associations were symbols in an associative
alphabet. This remains for future investigation.}.

\subsection {Examples of contextualization}

The structure of the following examples helps to show why context is so
important, and why it is important to aggregate clusters into concept-property
matroids and situation-context groups.  This is a set-theoretic
interpretation.  Contextual information is both interior
(explanatory of subjects) and exterior (causal and situational or
correlated). The following cases help to exemplify the issue:
\begin{itemize}
\item The doctor (in the role of surgeon) depends on surgical gloves in the context of (operating room,examining patients).
\item Bruce Wayne (in the role of Batman) is a member of super-heroes (in the role of nightclub).

\item Tidal is generalized by (is a kind of) company when (thinking about music, streaming music)
\item Tidal in the role of gravitational effect does NOT have the role of company.
\item Tidal is NOT caused by the moon in the context of tidal as a company.
\item Tidal has the role of an effect when thinking about physics, gravity.

\item Gravity depends on mass in the context of physics.
\item Gravity is the name of a book (a book called gravity) in the context of physics.

\item Temperature expresses very hot in the context of measuring server.
\item Hot contains very hot in context (all contexts).
\item Very hot contains `50 degrees C' in the context of measuring server.
\item Temperature expresses hot in the context of (measuring server, 306, datacentre, NYC).
\item Temperature expresses hot in the context of June.
\end{itemize}
These singular context names, while meaningful to readers with a broad
cultural knowledge, are too simple to allow meaningful discrimination during
machine processing. We need to understand how to extend context labels
for efficient addressing of semantic relevance.

\subsection{Context design and compound concepts}\label{intext}

Because of the recursive nature of concepts, the promise model of
scaled agency in \cite{spacetime2} suggests that it will be helpful to
distinguish between {\em interior} and {\em exterior} context of a
concept. Interior context refers to qualifying associations positioned
`behind' hubs, which help to support and qualify the interpretation of
a concept. Exterior context refers to the set of concepts that are
active in the recent train of thought of the total cognitive system,
as it interfaces with the world on the other end of its sensory
apparatus (see table \ref{intexttab}).  
\begin{table}[ht]
\small
\begin{tabular}{c|c|c}
CONTEXTS & \sc Interior & \sc Exterior\\
\hline
\sc Learning source / & Introspective & Observational\\
\sc awareness  & thinking & sensing\\
\hline
\sc Memory address /& Qualifying  & Relative\\
\sc retrieval & property & placement\\
\hline
\end{tabular}
\caption{\small Semantic scaling leads to a notion of interior and exterior at many levels.
  These distinctions play an important role in defining cognitive linguistic recursion during the
  learning and retrieval of stories.\label{intexttab}}
\end{table}

Interior memory context acts as a kind of `type' or `role' interpretation for
concepts that have been selected or deselected by an exterior {\em memory context}, whereas
interior {\em awareness context} represents the feedback of the system talking to itself
about past experiences and thoughts.
In all cases, the functional treatment of memory context may be handled by an
invariant algorithm, with great computational simplicity.
\begin{figure*}[ht]
\begin{alltt}
\footnotesize
 // What kinds of compound contexts can we expect? State of mind...
 // What are we trying to do?
 
 ContextCluster("doctor service");
 ContextCluster("patient appointment");
 ContextCluster("patient health service");
 ContextCluster("need to visit a doctor");
 ContextCluster("patient doctor registration"); 
 ContextCluster("identity authentication verification");
 
 // Compound (qualified) concepts

 RoleCluster("GP doctor","doctor", "general practitioner", "patient health service");
 RoleCluster("surgeon doctor","doctor", "surgeon", "patient health service");
 RoleCluster("patient appointment","appointment", "doctor patient", "patient health service");

 // Elementary bidirectional associative links

 Gr("doctor",a_promises,"doctor availability","patient doctor registration");
 Gr("health service access",a_depends,"patient authenticated","need to visit a doctor");
 Gr("health service access",a_depends,"patient appointment","need to visit a doctor");
 ....
\end{alltt}
\normalsize
\caption{\small Example programming statements leading to the formation of a
  graph from within a programming environment. Any system could, in
  principle, be instrumented from within to offer semantic
  hints for aggregation by a central `brain'. Helper functions
divide up and generate patterns, as illustrated in figures \ref{compounds2} and \ref{compounds1}.\label{API}}
\end{figure*}
Context is not just about the addition of descriptive words during
learning, but also about linking those compound states to all the
sub-concepts on which they depend too. By following this approach, the
average degree of connectivity in the graph increases {\em
  superlinearly}, making non-trivial stories increasingly likely,
while also providing switches by which to exclude pathways based on
later context during retrieval.

The recursion in the graph corresponds to a recursive linguistic structure
that can be illustrated as follows. Consider the following unary association,
containing subgraphs, labelled by their irreducible ST-type.
\begin{alltt}\it
\small
(the cat which (is black (4), eats fish (2)))
   \textbf{expresses}(4) (happy purr) in the context of (living room)
\end{alltt}
The parenthetic recursions are all {\em interior
  properties}, and the explanations of situation are {\em exterior} (boldface).
We end up with useful statements of the general structure:
\begin{alltt}\it
\small
(
 (concept name, interior recursive properties) 
  (\textbf{association type, alias})
 (concept name, interior properties) 
 in (exterior context)
)
\end{alltt}
This suggests that an API, something like the one shown in figure
\ref{API}, can help to automate the structural regularity during
the documentation of concepts and associations, i.e. during semantic learning.  Notice how, by using the
matroid hubs and compound concepts to reduce the dimensionality of
conceptual activations into a single concept, we also achieve an
economy of scale, with attendant computational simplification\footnote{It is
interesting to speculate whether this observation might help to
explain the common use of compound words and special terminology, like
acronyms, in human language.}.

\section{Retrieval of `knowledge'}

Once encoded, retrieval of associations is achieved by reparsing the
representation in figure \ref{prism}, subject to an independently
maintained context. Sensory context is a set of somewhat elementary
tokens, which activates certain concepts and associations. Exterior
context may be exchanged for slower, but more stable introspection, in
which derived concepts are fed back into the context by `thinking
about' a particular topic. Tokens can be followed individually or
activated in parallel to search for the consequences of a particular
concept in the current context.  Implementing this context channel
will be the most challenging aspect of building a cognitive knowledge
representation. Only the first stages will be addressed in this report.

\subsection{Conceptual pathways}

\begin{figure}[ht]
\begin{center}
\includegraphics[width=6.5cm]{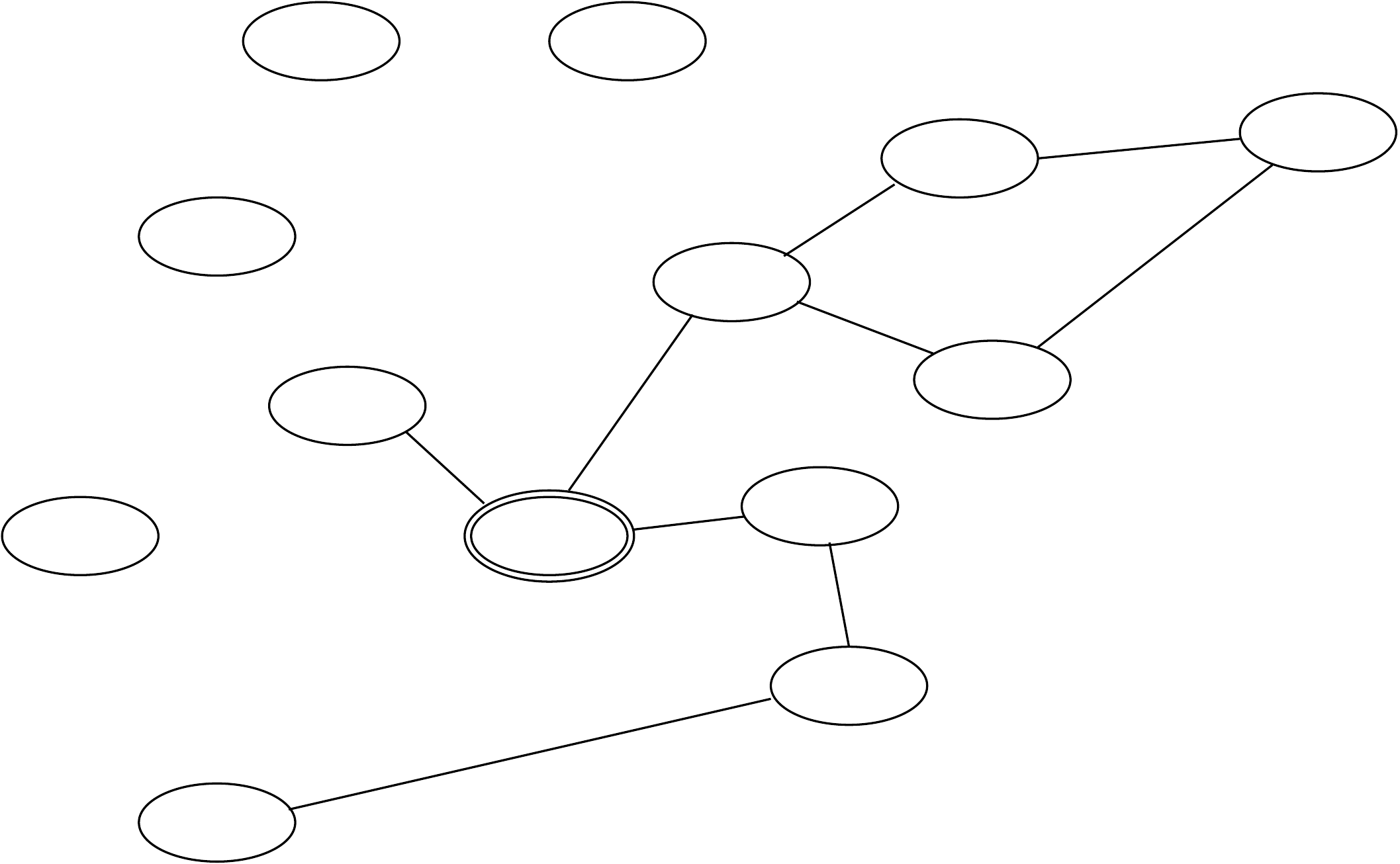}
\caption{\small Brainstorming around a start concept, without a clear end.
Reasoning may contain loops, which need to be handled somehow.\label{brainstorm}}
\end{center}
\end{figure}

Retrieval of explanations and associative reasoning can be open ended or bounded (see figure \ref{brainstorm});
we want to:
\begin{itemize}
\item Explain how to get from a starting concept and a final concept (bounded but not unique).
\item Explore where we can go from some a starting concept (unbounded brainstorming).
\item Explain the different ways one can get to a final concept (inverse brainstorming).
\end{itemize}
The first of these is somewhat analogous to a quantum path integral\cite{certainty}. Any one or several of the possible
pathways might be in play between concepts in a given context. For example, without sufficient
context to know the interpretations of `Oslo' and `America', both these unary associations
may be in play simultaneously during a reasoning process:
\begin{itemize}
\item Oslo is located in America
\item Oslo is NOT located in America.
\end{itemize}
In logic, this would be considered impossible, yet both of these
pathways may be equally relevant (we should avoid the expression
`true') if the context is unable to select only one. A knowledge
system cannot even know that these seem potentially contradictory
without a sufficient density of linkage to qualifying context.  We
must be quite careful not to over-constrain reasoning by projecting
human expectations onto a process too early. It is likely that
concepts such as true and false are emergent rather than prescriptive.

\subsection{Context in retrieval}\label{retrievecontext}

As we trace along a story, from concept to concept,
four semantic component sets may be associated with with each step in a story (see figure \ref{fournames}).
\begin{itemize}
\item The current concept in a story (a fixed boundary condition).
\item The concepts associated with the current concept (directed links to possibilities).
\item The current awareness context or `state of mind' of the observer (current relevance).
\item The awareness context at the time of learning (encoded in interior and exterior associations).
\end{itemize}
How these four sets activate one another is key to how stories may be
generated `deterministically'.  The current concept is a natural part
of the state of mind of an agent, and thus it belongs to the interior
cognitive context as much as any sensory inputs. The cognitive context
of an agent divides into interior and exterior parts. Interior context
is what the agent is thinking about, independently of its sensory
inputs. The exterior context is its sensory channels.  Thus some
context information is naturally inward-looking, or introspective,
while other information is outward-looking (grounding in the realities
at different times). It is interior awareness context that determines
the {\em relevance} of forward pathways, because state of mind is the
calibrator of intent. Exterior awareness context is about framing the
current experience, grounding it by experience or feeling. Thus, the
two cognitive `awareness' channels have different semantic
roles\cite{spacetime3}.

As context evolves, so does the focus of consciousness, or `what the
system is thinking about' i.e. the current concept or set of concepts
in a story frame.  Interior associative context plays the role of
functional type, and is entirely encoded in the recursive compounding
structures of the knowledge representation. So when we retrieve
knowledge, this kind of context is purely qualifying or explanatory of
the usage. The major links to concepts will be to the fully qualified
hub nodes, and the interior context will be hidden unless explicitly
examined, at any given level of recursion.

It is exterior associations that link concepts to one another in the
knowledge representation, so they determine which concepts lead to
other concepts. However, those exterior associations are judged for
relevance by comparing the context in which they were learned with the
current context or state of mind of the agent. This leads to a complex
causation: concepts overlap through association, and context overlaps
with pairs of concepts, and perhaps beyond.
We therefore have two channels evolving in parallel: a short term memory of
context and a long term memory of qualified concepts\cite{spacetime3}.

\subsection{Evolution of awareness context}

Measuring the relevance of forward pathways remains an unsolved
problem.  How should the current context evolve then as we traverse a
story. How quickly should context be forgotten?  When to increase the
scope of context and when to say that our evolving thoughts have
drifted off-topic is an highly subtle determination, which may not
lend itself to algorithmic or pattern based solution. It might well be
that relevance needs a bank of multiple pathways interfering across
the entire length of a story path.  When we apply software to the
problem, it is natural to treat each leg of a story as a Markov
process, each step independent of the last, with some activation
context adding transverse memory. This is probably too simplistic, but
for this work, that must be the limit of ambition.

As a first approximation, we may take the lexical overlap of the two
context sets (past and present) as an estimation of relevance. Using
this, we can sort stories link by link to judge the relevance of each
twist and turn. The relevance of the total story cannot be calculated
without a particular end-topic. That goes beyond the scope of these
notes.

When machine learning is applied to document processing, story fragments are treated as
patterns to be recognized, and the relevant forward pathways are determined
based on statistical ensemble support rather than semantic
relevance\cite{lifelong1,lifelong2}.

There are many issues with trying to match relevance based on context
in a small system.  Present context may not match the context at the
time of learning, because there are too few degrees of freedom for
coupling contexts together. It may turn out that we have to rely on
more universal invariant context markers like emotional states (fear,
happiness, etc) to play a role in activation and selection. Although these are
poorly understood, they may play a central role in joining together 
ideas that cannot easily be joined by explanations of semantics.

\begin{figure}[ht]
\begin{center}
\includegraphics[width=7.5cm]{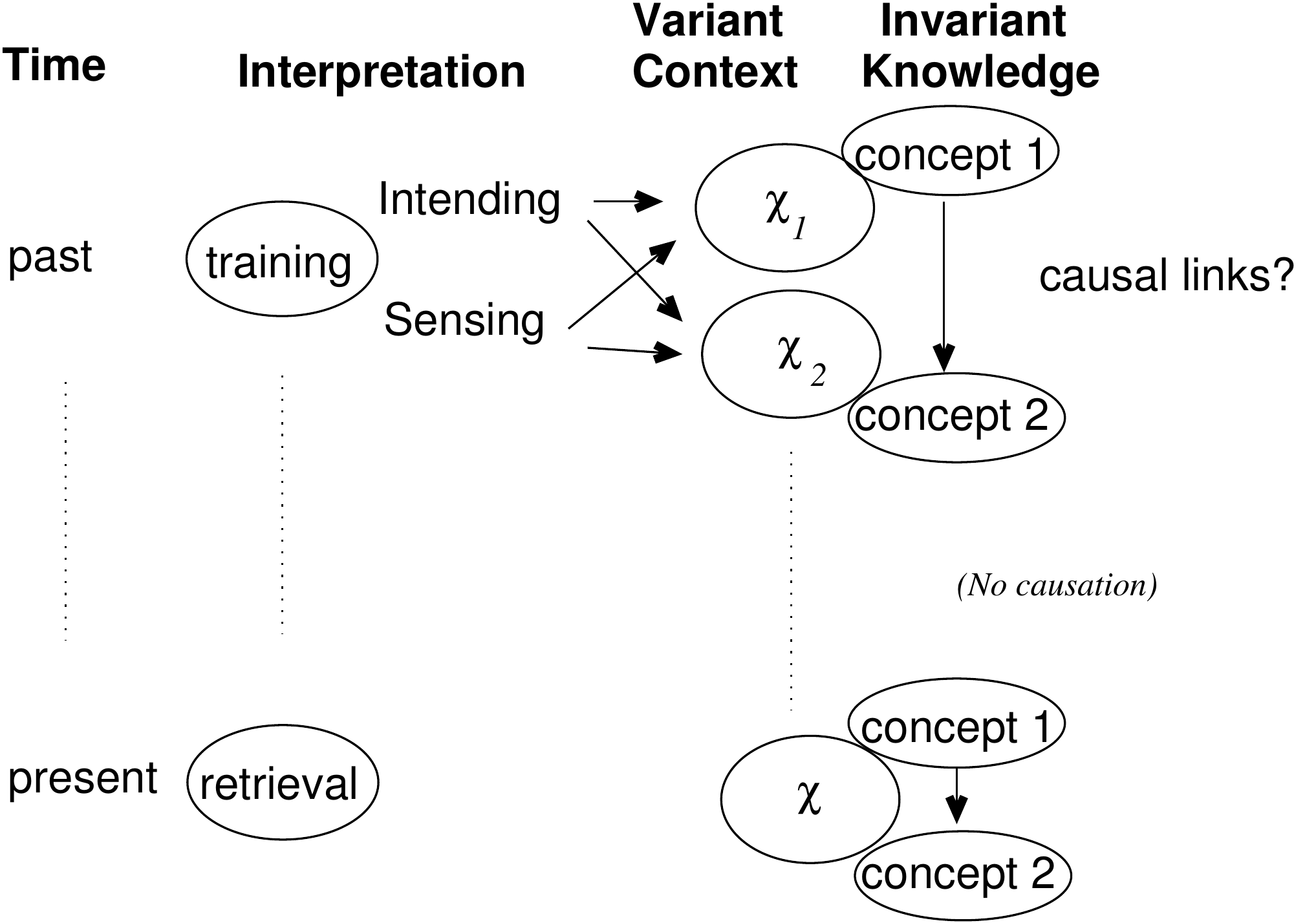}
\caption{\small The timescales associated with learning and retrieval
  of stories.  The links between concepts may or may not be causal
  over long separations, but context is always causally related to
  concepts over short timescales.\label{fournames}}
\end{center}
\end{figure}

As usual, it is important to remember that semantics are secondary to what
dynamics enable, thus causal propagation of meaning cannot be assumed across
different timescales (see figure \ref{fournames}).  We may expect the
timescales of learning and retrieval to become separated eventually,
with retrieval times much longer than training times, yet we must still
recall and modify stories that were learnt previously. This gives some clues about
the encoding. In particular it tells us that all causal information needs to be
encoded in exterior associations, not by relying on context to light up answers
directly, like a primary database key.

If excessive specificity can be the enemy of retrieval, then
specificity is important during the encoding phase of knowledge. For
example, it is not the general concept of birds that depends on the
concept of flight, but rather a fully qualified instance of
a bird. In other words, we have to be careful not to
confuse what is general with what is specific, at any level of
recursion, as this can lead to manifest errors of reasoning, such as
`all birds depend on flight'.  Matching of conceptual scales is yet
another problem that must be deferred for further study in future work.

\section{The experiments}

At the time of writing, a simple proof of concept implementation of
the foregoing cognitive system has been implemented. Only the simplest
level of interior/exterior context conditioning was used to activate
and screen paths, owing to the complexity of simulating such a
feedback system in software.  The initial goal was to illustrate the
principles.

\subsection{Data sources}

Four kinds of data source were imported into a semantic graph, in the hope of
first generating stories by brainstorming.  Even this simple first step has application for causal
analysis, qualitative hypothesis testing, etc. The data sources were:
\begin{itemize}
\item {\em Computer monitoring examples}: data from the Cellibrium
  CGNgine agent (a CFEngine `computer immunology' derivative, with
  embedded machine learning at the pattern level) were output in
  realtime, as in a production datacentre environment or IoT scenario.
  As an intentional system, the intended policy could also be encoded
  and connected to the activities of the agent, describing desired
  state. Thus data observations, intentionally measured from the
  actual state of the system, and converted into invariant forms,
  with an intended interpretation relative determined by policy goals
  and constraints. See figure \ref{story2}.

  Policy documentation files, describing intent, were incorporated, such as the
  intended meanings of software functions, with links to documentation,
  as well as the coding of particular policy rules.  The source code
  of the software agent could be scanned to extract error messages
  given in a particular context\footnote{Current programming languages
    do not make it easy to export the contextual intent behind code or
    user messages. In future, such an innovation will become essential
    to the scaled monitoring of a software society.}.  Domain
  knowledge about the design and intent of the software itself was
  included manually, as well as remarks about the world of computer
  datacentres, servers, and operating systems.

\item {\em Software system example}: Data scanned from the nesting
  structure and dependencies of software, packaged in process
  containers, file bundles, and linked binaries, probed using intended
  composition rules (source: Makefiles, container specifications,
  package compositions, and binary linker data from {\tt ldd}, etc).
  See figures \ref{story3} and \ref{story4}.

\item {\em Doctor online registration wizard}: Domain knowledge from
  an imaginary online scenario, where a patient is trying to register
  for a doctors appointment in a new city. Concepts and associations
  are input directly from a wizard overview of knowledge of intent
  about the smart distributed service, and the software it uses,
  its dependencies, and workflows etc., (source: authors of the
  system). See figure \ref{doctordoctor}.

\item {\em Big picture semantic index example}: Domain knowledge contributed from
possibly multiple sources giving a cognitive overview of an entire environment
or experience. See figure \ref{story5}.
\end{itemize}

\subsection{Machine learning}

Machine learning appears at two distinct levels:
\begin{enumerate}
\item The sensory data
collectors sample the exterior sources at some (relatively frequent)
timescale, in accordance with Nyquist's theorem.  Each collector, with
its independent semantics, transmutes sample data into a separate
invariant representation and learns its significant patterns and
states (for monitoring, see \cite{burgessC14}). From this Bayesian
statistical description, a lexical name representation is selected
from a small set invariant states.  
\item Associations are updated on a
regular timescale, and each associative link acts as an independent
Bayesian learning process (i.e. a Bayesian network), giving the links
in the semantic network weights relating to their importance and
frequency of visitation. This applies at each recursive scale of concept
aggregation.
\end{enumerate}
The intentional aspects of any system, which was designed or evolved
for a niche purpose, are central to the encoding of its semantics.  If
such information is not captured at the source, it is simply lost.
Users of the software may later reinterpret this intent, in the light
of current assumptions, but the intended meaning ultimately originates
from a source which matches two promises: the `promise/intent offered'
and `promise/intent accepted' (like lock and key).  The receiver
always has the last word in interpretation however\cite{promisebook}.

\subsection{Software implementation}

Prototype code for creating and parsing the knowledge representation,
is shared under the Cellibrium\cite{cellibrium} project. This employs
the simplest possible proof of concept, avoiding programming concerns
in favour of pedagogy.  Concepts are represented as directory names,
each containing subdirectories for the 4+1 association types; then, in
turn, a file of association annotations, including timestamps and
weights for Bayesian updating. Although filesystem limitations make
this approach unsuitable in the long run, it made the coding of a
prototype pleasantly free of dependencies and obscure APIs. Everything
could be handled with fast, efficient POSIX system calls\footnote{It
  would seem as though a graph database would be the ideal candidate
  for representing such data, but thus far, the input and query
  languages for such databases are clumsy and ill-suited to this kind
  of structure.}.
All concepts have directories under some root node, which contain subdirectories
corresponding to $\pm$ the four spacetime association types. Under these lie
files, which document the concepts that may be reached by following that type of
association:
\begin{alltt}
root/
root/StartConcept/
root/StartConcept/1-4/
root/StartConcept/1-4/NextConcepts*
\end{alltt}
The `next concepts' are files containing the specific annotations,
timestamps, and learning weights of the associations between the starting concept
and the chosen next concept.

To frame context, a separate control channel is used, mainly for
implementation convenience. The context association is a spacetime
containment type (ST3), but its functional role is different to other
groupings.  Contexts work in exactly the same way as other concepts,
with composites and sub-clusters.  However, contexts arise {\em ad
  hoc} and thus their clusters do not always have lasting meaning as
composite concepts. As running catalogues of what is currently going
on, somewhat analogous to a browser history, so relevant words
pertaining to the agent state were simply strung together in no
particular order.  This is analogous to the original CFEngine
classes\cite{burgessC1}, where context was a collection of class
strings that were probed from the environment by specialized software
sensors.  Combinations of these classes could be constructed as
quasi-logical expressions (acting as hubs) to activate certain
concepts (in this case configurations). The basic atoms were drawn
from system variables.  

In more general cases, at a higher level of
abstraction, context must include the `state of mind' of the cognitive
system over the recent past (again, think of browser history), which
encapsulates intent as well as environmental impulses.  These will be
expressible in terms of higher level concepts\footnote{Framing and
  limiting a context is a very hard pedagogical problem. As a semantic
  spacetime, a knowledge representation is somewhat similar to a
  higher dimensional field theory with an ultraviolet cutoff, as
  information is discrete and finite. A story is like a transition
  function, or $S$-matrix element, whose vertex rules are based on the
  causal associations and their propagation rules.}

\begin{figure*}[ht]
\begin{center}
\includegraphics[width=12.5cm]{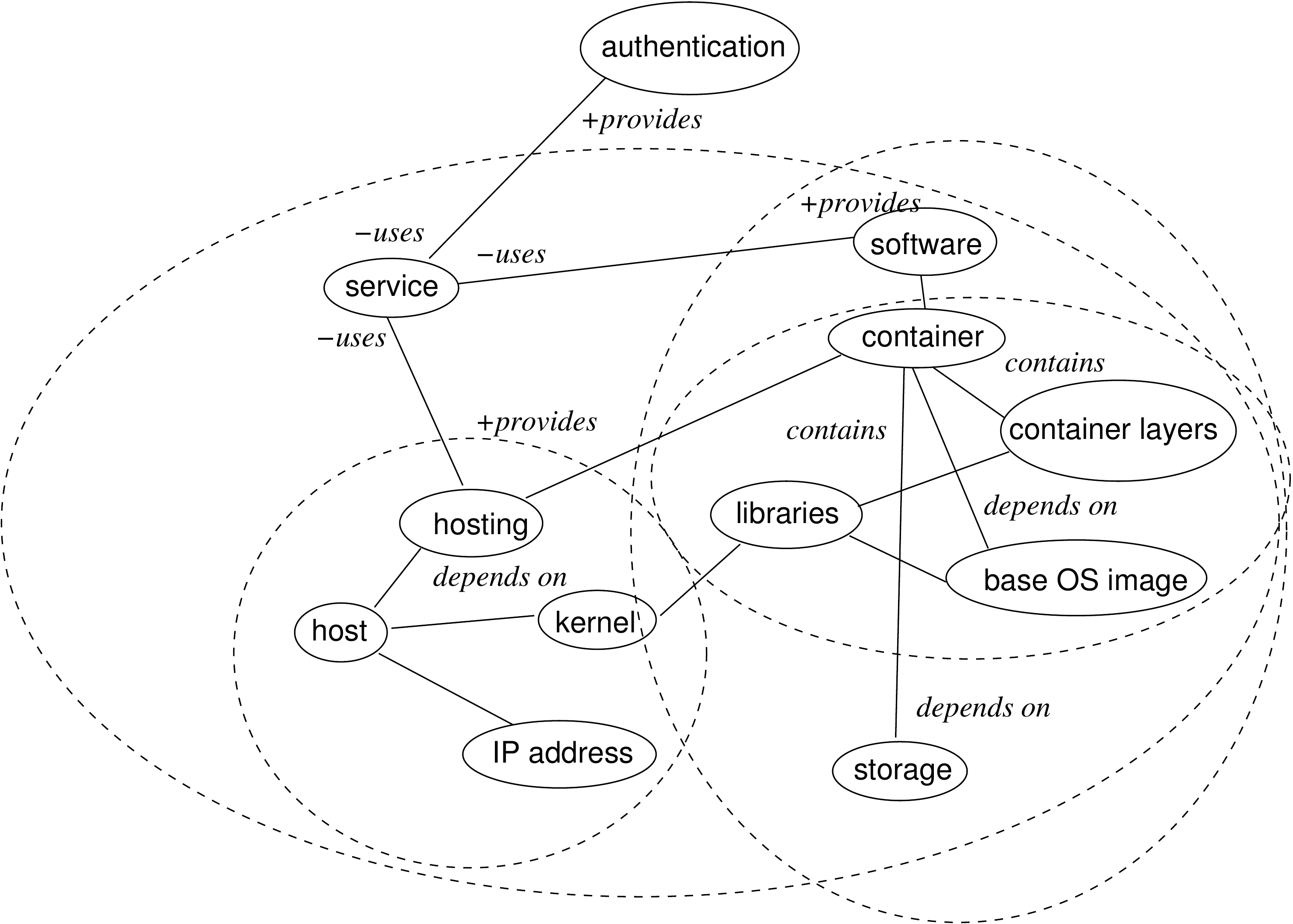}
\caption{\small Software associations used to propose a template for scanning
software services. From this model template, a distributed software crawler 
could be written in a user-secure manner to extract operational knowledge
about deployed software. This cluster of concepts illustrates how one
can build a template for a core ontology within a small domain.\label{microservice}}
\end{center}
\end{figure*}

\section{Results}

\subsection{Assessment}
Assessing the success of the prototype `brainstorming' experiments is
not an easy matter, given the qualitative nature of the results.  This
is not so much a weakness of the experiment, as the nature of the
problem.  Although it is possible to manufacture artificial metrics by the
collection of numerical data, e.g. about numbers of proposed stories,
or conceptual overlap with context along a path, etc, these measures
are themselves {\em ad hoc} and largely unhelpful, as they do not
correlate easily with the subjective sense of `a ha!' one has when
finding a helpful outcome.  So, we are left with a kind of `Turing test'
assessment approach, i.e. `it looks ok' to a human. Later, once one
applies the approach to a specific problem, where target goals can be
incorporated from the beginning, we should be able to do better in
establishing consistent criteria. I shall not address this further in
this report.

\subsection{Data}

Data sets used consisted of hundreds to thousands of concept tuples,
collected from a variety of sources.  These numbers are quite small
for a realistic application, but suffice to test some comparative
aspects of the approach.  Association data were generated by
annotating system monitoring in realtime, from software outputs,
documentation scans, and by manual knowledge documentation of hints.

In early trials, realtime monitoring data were fed into a knowledge
representation in a way that retained a lot of specific information,
to see if this might stabilize into a comprehensible emergent pattern.
However, this experiment quickly revealed that the potential burden of
remembering the detailed semantics of every moment.  

If each new measurement set leads to a different context, i.e. making
time itself the generator of context, the number of concepts (or
amount of data) would be prohibitively large (not for a computer
to remember, but for any human or algorithm to make sense of). One cannot
escape the fact that our human brains have builtin limitations.

Since we are dealing with semantic interpretations, we cannot simply
say `big data', and treat every knowledge problem as a forensic
investigation, because we cannot guarantee that a focused set of
concepts will emerge from the data. That is a different problem than
having an expert on hand who is already au fait with the necessary
knowledge.  Thus, we must look for invariant representations of big
data, cumulatively over time, that may be interpreted by only a {\em
  small} number of (possibly composite) concepts, for the sake of
comprehension.  Fortunately, studies have shown that meaningful
patterns from system monitoring behaviours are few in number, compared
to a bulk of non-meaningful noise\cite{burgessC8}.

\subsection{Structure}

Throughout early experiments, the output of a brainstorming episode,
based on a given starting concept, tended to lead to either no stories or
too many stories. The reasons for this depended on both the structure of
data and the selection criteria.  Initial attempts to find quality
stories were hampered simply by an absence of data.  The first trials,
using data types to represent contexts, were relatively unsuccessful.
The input of knowledge took a huge manual effort to curate, both in
terms of syntactic and semantic burdens placed on the author; then,
even if someone could be persuaded to undertake encoding, it would be
fragile and under-connected. Very little unexpected emergence was
possible, as the typing made data overconstrained.  The sparseness and
fragility of the typed structure meant that concepts could not easily
be discovered by free association, and most searches resulted in no
results.

Topic map inspired structures\cite{topicmaps} created mainly led to
trivial stories, so the project was set aside for almost 4
years.  Following this hiatus, the models based on topic maps were
abandoned\cite{burgesskm}, and ideas from the promise theoretic
approach \cite{spacetime1} were revived.  

The removal of `types' as discriminators, and their replacement with
context-free lexical terms, led to a large increase in the density of
stories. It is difficult to put numerical measures to the improvement,
because a radical restructuring of many aspects was involved in this
shift, and the results are very dependent on the particular
characteristics of example data, which were not preserved over the
intervening years between the difference, although an attempt was made
to reconstruct similar data. At best one could say that there was a
qualitative improvement in the results.

The earliest attempts at discovering emergent connections, were implemented
by my collaborator Alva Couch\cite{stories,inferences}, and used a shortest path
approach to selecting a unique `route' between two concepts: initial
and final boundary conditions. This is slightly different from full
brainstorming, because it is already more constrained by having a
definite target concept.

Ultimately, I abandoned the shortest path approach as it creates a
tension between quantitative metrics and dynamical selection criteria
rather than semantic relationships. The idea that a single path, based
on its length, could be a correct answer to the exclusion of others
does not well motivated; although, it is plausible that, given a set
of semantically valid paths from $A$ to $B$, the shortest one might
have interesting qualities.  The advantages of fixing boundary
conditions at the start and the end of a story were compelling, but it
is not always so clear what we are looking for when exploring causes
and interpretations.

\subsection{Computational complexity}

The derivation of the four irreducible spacetime association types
brought a significant simplification to the algorithm for exploring
locales.  Initially a linear search was needed to identify matching
concepts from a number of type `namespaces'. Then associations could
be followed, but the question of whether the associations could
propagate or not was not encoded directly into the association,
requiring another lookup, or so-called inference rule. The possible
complexity was of the order of the product of the length of the
inference table with the number of outgoing associations (the
out-degree of the graph).  Thus the pre-classification according to
the spacetime `compass types' led to a reduction in the order of
magnitude of the computational complexity by eliminating the need for
the inference table. The propagation of these types in inferences is
defined by the spacetime classifications themselves\cite{spacetime3},
and no matching is required.

A further reduction in the degree of `grammatical complexity' may come
from the aggregation of multiple concepts into aggregate classes. If
associations can be made to larger umbrella concepts, rather than
concepts, which are too specific, then the cost of recursion can be
spared. Thus, what may seem to be a context-free expression,
linguistically, may be sufficiently represented by a regular language
expression. See \cite{spacetime1,spacetime3} for a discussion of the
grammatical complexity of spacetime associations.

The addition of context, for either positive or negative matching, is
the most intensive computation remaining; however, this observation
seems critical to establishing semantic relevance of stories.
Context, as modelled in the original CFEngine
representation\cite{burgessC1,tiny}, and its derived Cellibrium
implementation\cite{cellibrium}, is a non-ordered linear list composed
of many atomic characterizations.  Each evaluation of context requires
parsing the entire list for matches with a similar non-ordered
expression composed of the same atoms. The complexity could be of the
order of the product of the current context lookup and the learned
context lookup, for each association.  Supposing that each could be of
logarithmic order, by hashing, this is much better than a model
without the encoding of spacetime types.  The size of a context cannot
be pre-determined, because it fluctuates over short times, and grows
during long-term learning.

Each link in a story has a complexity cost, associated with
the relevance of the link. This was measured by the size of the overlap
between current context and learned context (analogous to the mutual
information transfer\cite{shannon1,cover1}). In the CFEngine model,
context was defined as a `logical boolean expression', which enabled
contextual relevance to be reduced even further to a single expression
evaluation, whose overlap was generally quite small, i.e. only the
length of a very specific boolean expression\cite{tiny}.  There are many ways to
optimize such contextual matches, using aggregate classifications and
caching, largely because each execution of a CFEngine context is a
frozen moment in time.  The nature of a cognitive system, on the other
hand, is that context is continuously changing, so it will always be
more expensive to compute.

\begin{figure*}[ht]
\footnotesize
\begin{alltt}
\it
host$ {\bf ./stories -s "Cannot mix CIDR notation with xxx-yyy range notation *" -c "software fault" -t 2}
0:follows)  "Cannot mix CIDR notation with xxx-yyy range notation *" may be caused by "network policy" 
           (in the context of errors and faults)
1:apprxnr)    "network policy" contains "interfaces configuration" (in the context of software)
0:follows)  "Cannot mix CIDR notation with xxx-yyy range notation *" may be caused by "FuzzySetMatch" 
            (in the context of errors and faults)
1:preceds)    "FuzzySetMatch" may be used by "iprange" (in the context of software)
2:follows)      "iprange" depends on "network pattern matching" (in the context of software)
0:follows)  "Cannot mix CIDR notation with xxx-yyy range notation *" may be caused by 
           "incorrect use of network pattern matching" (in the context of errors and faults)
1:hasprop)    "incorrect use of network pattern matching" is related to "network pattern matching" (in the context of errors and faults)
2:preceds)      "network pattern matching" partly determines "iprange" (in the context of software)
3:follows)        "iprange" may use "FuzzySetMatch" (in the context of software)
3:apprxnr)        "iprange" belongs to or is part of "CGNgine class function" (in the context of software)
4:apprxnr)          "CGNgine class function" belongs to or is part of "system policy" (in the context of software)
4:apprxnr)          "CGNgine class function" belongs to or is part of "CGNgine functions" (in the context of software)
5:apprxnr)            "CGNgine functions" belongs to or is part of "CGNgine policy language" (in the context of software)
3:cntains)        "iprange" has the role of and expresses an attribute "host classifier" (in the context of software software)
4:expr-by)          "host classifier" also known as "class or context label" (in the context of system monitoring)
\end{alltt}
\caption{A brainstorming rooted in an error message generated in a
  CGNgine system log. This story frames the possible cause(s) of the
  initial error message anchor point nicely, and pin-points the
  offending function `iprange'.\label{story2}}
\end{figure*}

\subsection{Sensemaking of stories from structure}

Having enough brainstorming paths to analyze, in the output of a
search, was an advantage of the new spacetime approach to knowledge
representation.  Two problems remained however, whose remnants may
still be seen in the figures:
\begin{itemize}
\item Stories were often unreadable and bizarre, as even valid lexical phrases coughed
up along pathways of the graph had little meaning without the original intent of the
associations and names preserved and represented. The loss of intent seemed to lie
at the heart of this. Intent acts itself as a form of context.

\item Often, stories did not terminate, without an artificial length
  cutoff, in spite of loop detection methods on a per-story basis, as
  the simplistic recursion allowed stories to contain one another as
  sub-parts by the free association.  This suggested that success may
  lie in the use of context to constrain stories and bring focus to
  the results.
\end{itemize}
Later, when repeated visitation of a concept was forbidden across all
stories (as a global constraint), the stories did become finite;
however, now an argument could be made that the search was now
overconstrained, and ad hoc choices might exclude certain combinatoric
stories from being visited at all\footnote{The original approach used by CFEngine
  was to simulate the `neuronal dead-time' into concepts, eliminating
  loops by virtue of concepts having fired already\cite{lisa97113}.
  This worked in CFEngine, because of its simplistic linear processing
  approach.  In a network with multidimensional causation, the
  approach could easily prevent important answers from emerging rather
  than protect against repetition.}.  

In fact, the non-termination of stories is another indication that the
graph is well connected, even with the small data sets used, but they
were also an artefact of the open-endedness of the brainstorming
process. Without a clear goal either in terms of context or specific
concept, there is no deterministic criterion for running out of
combinatoric options.

The recursive scaling of ideas anchors ideas to context in a more
powerful way of attaching intended context than typing. This is
basically analogous to a transition to a schemaless database.
After the introduction of the recursive hub structure, and the
automation API, in figure \ref{API}, the story density was largely
unchanged, but the readability of the stories was greatly improved,
allowing the kinds of outputs shown in the figures. The structure led
naturally to the separation of context into interior and exterior memory
channels as mentioned in section \ref{intext}.  The interior memory context,
in particular, brought immediate comprehension, even when the strict
linguistic form was contrived. This, of course, is more a testament to
human linguistic acumen than to the skill of a very simple algorithm,
but it indicates that the essence of language is captured by a
recursive identification of irreducible types.

\subsection{Outputs}

Figures \ref{story2}-\ref{story5} show typical samples of outputs from
a brainstorming. The actual outputs are much longer than these. This
print format is not well suited to show the data. The left hand margin
shows the spacetime association types (ST1-ST4) followed to make the
association, and the indentation shows recursion depth along a single
path.

\begin{figure*}[ht]
\footnotesize
\begin{alltt}
host$ ./stories -s "microservice"
\it
0:apprxnr)  "microservice" is approximately "service" (in the context of software ops operations develop write)
1:follows)    "service" may use "user authentication" (in the context of software ops operations develop write)
2:hasprop)      "user authentication" has the role of "authentication" (in the context of software ops operations develop write)
       and also note "authentication" is a role fulfilled by "service authentication" (in the context of software ops operations develop write)
1:follows)    "service" may use "service authentication" (in the context of software ops operations develop write)
2:hasprop)      "service authentication" has the role of "authentication" (in the context of software ops operations develop write)
       and also note "authentication" is a role fulfilled by "user authentication" (in the context of software ops operations develop write)
1:follows)    "service" depends on "software SOFTWARENAME" (in the context of software ops operations develop write)
2:follows)      "software SOFTWARENAME" depends on "container CONTAINERNAME" (in the context of software ops operations develop write)
3:follows)        "container CONTAINERNAME" depends on "package PACKAGENAME2 VERSION2" (in the context of software ops operations develop write)
4:hasprop)          "package PACKAGENAME2 VERSION2" has the role of "package" (in the context of software ops operations develop write)
           and also note "package" is a role fulfilled by "package PACKAGENAME1 VERSION1" (in the context of software ops operations develop write)
           and also note "package" is a role fulfilled by "package PACKAGENAME3 VERSION3" (in the context of software ops operations develop write)
4:follows)          "package PACKAGENAME2 VERSION2" depends on "library LIB1" (in the context of software ops operations develop write)
5:preceds)            "library LIB1" partly determines "package PACKAGENAME1 VERSION1" (in the context of software ops operations develop write)
6:hasprop)              "package PACKAGENAME1 VERSION1" has the role of "package" (in the context of software ops operations develop write)
               and also note "package" is a role fulfilled by "package PACKAGENAME3 VERSION3" (in the context of software ops operations develop write)
5:preceds)            "library LIB1" partly determines "package PACKAGENAME3 VERSION3" (in the context of software ops operations develop write)
6:hasprop)              "package PACKAGENAME3 VERSION3" has the role of "package" (in the context of software ops operations develop write)
               and also note "package" is a role fulfilled by "package PACKAGENAME1 VERSION1" (in the context of software ops operations develop write)
1:follows)    "service" depends on "hosting" (in the context of software ops operations)
2:expr-by)      "hosting" is a role fulfilled by "hosting CONTAINERNAME INSTANCE" (in the context of software ops operations develop write)
2:follows)      "hosting" depends on "host HOSTNAME" (in the context of execute run software ops operations)
\end{alltt}
\caption{Knowledge scanned from a distributed application (simulated) shows the structure of the service, its software and
dependencies, from containers to operating system package names. The litany of components and service dependencies might be used
for troubleshooting or risk/vulnerability/impact analysis.\label{story3}}
\end{figure*}

In figure \ref{story2}, the data are taken from the CGNgine
software, using several data transmutors to scan both the source code,
and the system policy. The occurrence of an obscure error message in a system
output can immediately be tied to the likely sources and points of expression,
allowing a user to trace the source all the way back to their intended policy,
and through the software layers. 

Figure \ref{story3} shows data acquired by a software build pipeline. It shows how 
deployed software depends on its containerized packaging, software dependencies, and 
host operating system base, all the way down to the location of the deployment. 
This chain of dependencies may be of use to developers, operations, and risk analysis
personnel, to name a few. Further analysis based on the specific semantics
could be automated further. Simply gathering this information into a single place,
in a readable format is no small task.
\begin{figure*}[ht]
\footnotesize
\begin{alltt}
\it
0:follows)  "application cgn-agent" depends on "cgn-agent" (in the context of host configuration maintenance agent CGNgine)
1:follows)    "cgn-agent" depends on "libz.so.1" (in the context of host application software dependencies security)
2:follows)      "libz.so.1" depends on "libc.so.6" (in the context of host application software dependencies security)
3:follows)        "libc.so.6" depends on "!lib64!ld-linux-x86-64.so.2" (in the context of host application software dependencies security)
4:preceds)          "!lib64!ld-linux-x86-64.so.2" partly determines "libkrb5support.so.0" (in the context of host application software dependencies security)
5:follows)            "libkrb5support.so.0" depends on "libpcre.so.1" (in the context of host application software dependencies security)
6:follows)              "libpcre.so.1" depends on "libpthread.so.0" (in the context of host application software dependencies security)
7:preceds)                "libpthread.so.0" partly determines "libdbus-1.so.3" (in the context of host application software dependencies security)
8:preceds)                  "libdbus-1.so.3" partly determines "libavahi-client.so.3" (in the context of host application software dependencies security)
9:follows)                    "libavahi-client.so.3" depends on "libdl.so.2" (in the context of host application software dependencies security)
10:preceds)                      "libdl.so.2" partly determines "libcrypto.so.1.0.0" (in the context of host application software dependencies security)
10:preceds)                      "libdl.so.2" partly determines "libpq.so.5" (in the context of host application software dependencies security)
     ++ more ....
10:preceds)                      "libdl.so.2" partly determines "libp11-kit.so.0" (in the context of host application software dependencies security)
10:preceds)                      "libdl.so.2" partly determines "libgssapi_krb5.so.2" (in the context of host application software dependencies security)
     ++ more ....
10:preceds)                      "libdl.so.2" partly determines "libgcrypt.so.20" (in the context of host application software dependencies security)
     ++ more ....
9:follows)                    "libavahi-client.so.3" depends on "libavahi-common.so.3" (in the context of host application software dependencies security)
10:preceds)                      "libavahi-common.so.3" partly determines "libvirt.so.0" (in the context of host application software dependencies security)
     ++ more ....
9:preceds)                    "libavahi-client.so.3" partly determines "libvirt.so.0" (in the context of host application software dependencies security)
10:follows)                      "libvirt.so.0" depends on "libcrypto.so.1.0.0" (in the context of host application software dependencies security)
     ++ more ....
10:follows)                      "libvirt.so.0" depends on "libidn.so.11" (in the context of host application software dependencies security)
10:follows)                      "libvirt.so.0" depends on "libudev.so.1" (in the context of host application software dependencies security)
10:follows)                      "libvirt.so.0" depends on "libnl-3.so.200" (in the context of host application software dependencies security)
10:follows)                      "libvirt.so.0" depends on "libp11-kit.so.0" (in the context of host application software dependencies security)
     ++ more ....
8:preceds)                  "libdbus-1.so.3" partly determines "libvirt.so.0" (in the context of host application software dependencies security)
9:follows)                    "libvirt.so.0" depends on "libcrypto.so.1.0.0" (in the context of host application software dependencies security)
10:follows)                      "libcrypto.so.1.0.0" depends on "libdl.so.2" (in the context of host application software dependencies security)
     ++ more ....
10:preceds)                      "libcrypto.so.1.0.0" partly determines "libpq.so.5" (in the context of host application software dependencies security)
     ++ more ....
\end{alltt}
\caption{Dependency relationships obtained by a simple scan of a process container are a hopeless task for a human
to trace. The level of interdependency is very large. The ability to trace dependencies across software deployed
in multiple distributed locations could be central to addressing transmission of faults and attacks. We see that context strings can become quite long. This is both expected and welcome, as it increases the possibility to filter and conjoin relationships based on specific circumstances, moving away
the the kind of general brain-dumps shown here.\label{story4}}
\end{figure*}

Figure \ref{story4} shows a more straightforward recursive
decomposition of library dependencies discovered by using the Linux
{\tt ldd} command. The deeply nested and extensive list of cross
dependencies is quite impossible for any human to comprehend, but is
straightforward to encode for analysis.

Figure \ref{doctordoctor} shows a facsimile of the distributed system
used by patients to register with a new doctor in Norway, as curated
from the perspective of a domain expert. The procedure involves going
to a website, but first one must have a number of credentials from
third party sites, totally unrelated to the public service. This kind
of constellation of independent services, integrated into a meaningful
whole would be handled by some kind of wizard software if it were a
task for a single computer. In the web era, there is no single agency
with the responsibility to integrate this information. A cognitive
system, which monitors services in a smart city, could do this, for instance.

Figure \ref{story5} shows some more playful examples of domain
knowledge, concerning the purpose of different services, in different
contexts.  A lexical token like `tidal' might refer to many different
things in different contexts. Here the brainstorming is able to
distinguish these cases using the recursive structure of the matroid
hubs.  In a more sophisticated rendition of the brain dump, one would
be able to select specific interpretations from these structures
without risk of conflict.  Whether the source of the semantics is
manual curation or trained natural language processing, it does alter
the key benefit of the approach used here: namely that it is orders of
magnitude cheaper to build and to use than an approach based on
repeated use of `big data'. Such approaches have been used to discover
vector congruences and emergent functional semantics, but only with
extensive training \cite{wordreg1,wordreg2,wordreg3,MahadevanC15}.
Such approaches could still be used as inputs to the present approach,
in the absence of something more practical.

\subsection{Inadequate context}

During the development of this approach, the notion of context was
subject to a lot of confusion and misdirection. The way logic-based
approaches to semantic modelling have tried to use context (as data
types) turned out to be quite wrong: there exists no objective version
of knowledge that can determine a fixed type-hierarchy or preferred
ontology. Knowledge is very much in the eye of the beholder, and every
observer forms an individual ontology\cite{burgesskm}.  Context is not
an objective categorization, it is a running state of mind (like a
browser search history that records our intent).  As long as we try to
treat a knowledge representation as an objective static boundary
condition on knowledge, rather than as a private subjective
interpretation, we will run into difficulties.

In this present approach, motivated by spacetime classification of
semantics, and which thus far ignores any preferred ordering by link
weights, semantic stability is encouraged simply by a structural
aggregation of concepts, analogous to human language. So, even if
there are random variations, using promises like `A is near B'
effectively allows us to collapse disconnected or multiple stories
into a single broader line story. What is important to reasoning is
that there lies a causal set of paths, that may be integrated with the
specific boundary conditions from start to end\footnote{This is
  tantalizingly reminiscent of a quantum path integral, for reasons
  the reader may ponder without suggestion from me.}.

\begin{figure*}[ht]
\footnotesize
\begin{alltt}
host$ ./stories -t 2 -s "doctor service" |more
Found story subject: "doctor service"
Stories of type 2 only
\it
0:follows)  "doctor service" depends on "patient appointment" (in the context of need to see a doctor)
1:follows)    "patient appointment" depends on "patient doctor register binding" (in the context of need to see a doctor)
2:follows)      "patient doctor register binding" depends on "doctor patient binding service" (in the context of patient doctor registration)
3:follows)        "doctor patient binding service" depends on "patient authenticated" (in the context of patient doctor registration)
4:follows)          "patient authenticated" depends on "identity credentials" (in the context of identity authentication verification)
5:follows)            "identity credentials" may originate from "https:!!url1!form!element2" (in the context of identity authentication verification)
5:follows)            "identity credentials" uses step 1 and may originate from "https:!!url1!form!element1" (in the context of identity authentication verification i
dentity authentication verification)
5:follows)            "identity credentials" uses step 1 "https:!!url2!form!element2" (in the context of identity authentication verification)
5:preceds)            "identity credentials" are required for and partly determines "doctor authenticated" (in the context of identity authentication verification ide
ntity authentication verification)
6:preceds)              "doctor authenticated" is required for and partly determines "accepted doctor patient binding" (in the context of patient doctor registration 
patient doctor registration)
7:follows)                "accepted doctor patient binding" depends on "doctor authorized" (in the context of patient doctor registration)
7:follows)                "accepted doctor patient binding" depends on "doctor" (in the context of patient doctor registration)
7:follows)                "accepted doctor patient binding" depends on "patient" (in the context of patient doctor registration)
4:preceds)          "patient authenticated" partly determines "have public health service access" (in the context of need to visit a doctor)
5:follows)            "have public health service access" depends on "public health service available" (in the context of need to visit a doctor)
6:follows)              "public health service available" depends on "general practictioner doctor available" (in the context of need to visit a doctor)
6:follows)              "public health service available" depends on "open for business" (in the context of need to visit a doctor)
6:preceds)              "public health service available" partly determines "public health service" (in the context of need to visit a doctor)
7:follows)                "public health service" depends on "patient uses appointment" (in the context of need to visit a doctor)
7:follows)                "public health service" depends on "doctor availability" (in the context of need to visit a doctor)
3:follows)        "doctor patient binding service" depends on "doctor authenticated" (in the context of patient doctor registration)
4:follows)          "doctor authenticated" depends on "identity credentials" (in the context of identity authentication verification)
5:follows)            "identity credentials" may originate from "https:!!url1!form!element2" (in the context of identity authentication verification)
5:follows)            "identity credentials" uses step 1 and may originate from "https:!!url1!form!element1" (in the context of identity authentication verification i
dentity authentication verification)

Total independent outcomes/paths = 28
Estimated relevance of outcomes 0/0

\end{alltt}
\caption{\small Cooperating distributed (micro)services may have no cohesive storyline to connect them, without aggregating the intended relationships between them. Intentional (promise oriented) documentation allows us to generate a kind of documentation `wizard' in realtime from processes changing in realtime.
Causal links show the representation of a complicated distributed system, which could be the basis of a wizard for integrating apparently unrelated parts in a single view.\label{doctordoctor}}
\end{figure*}

\begin{figure*}[ht]
\footnotesize
\begin{alltt}
host$ ./stories -s "microservice"

\it
0:hasprop)  "microservice" has the role of "application" (in the context of software ops operations develop write)
   and also note "application" is a role fulfilled by "application cgn-agent" (in the context of host application software dependencies security)
   and also note "application" is a role fulfilled by "application service" (in the context of software ops operations develop write)
1:apprxnr)    "application" has instance "tidal" (in the context of online services)
2:follows)      "tidal" depends on and NOT depends on "facebook" (in the context of online services applications physics gravity orbit moon satellite distortion)
3:expr-by)        "facebook" offers service "facebook authentication" (in the context of logging in)
3:cntaind)        "facebook" offers service "facebook authentication" (in the context of logging in)
2:follows)      "tidal" is authenticated by "facebook authentication" (in the context of online services applications)
3:hasprop)        "facebook authentication" is a service offered by "facebook" (in the context of logging in)
3:cntains)        "facebook authentication" is a service offered by "facebook" (in the context of logging in)
     and also note "tidal" is generalized by "gravitational effect" (in the context of physics gravity orbit moon satellite distortion)
2:cntains)      "tidal" is generalized by "gravitational effect" (in the context of physics gravity orbit moon satellite distortion)
3:follows)        "gravitational effect" depends on "distance" (in the context of physics)
3:follows)        "gravitational effect" depends on "mass" (in the context of physics)
1:cntaind)    "application" has instance "tidal" (in the context of )
2:follows)      "tidal" depends on and NOT depends on "facebook" (in the context of online services applications physics gravity orbit moon satellite distortion)
3:expr-by)        "facebook" offers service "facebook authentication" (in the context of logging in)
3:cntaind)        "facebook" offers service "facebook authentication" (in the context of logging in)
2:follows)      "tidal" is authenticated by "facebook authentication" (in the context of online services applications)
3:hasprop)        "facebook authentication" is a service offered by "facebook" (in the context of logging in)
3:cntains)        "facebook authentication" is a service offered by "facebook" (in the context of logging in)
2:apprxnr)      "tidal" is generalized by "gravitational effect" (in the context of physics gravity orbit moon satellite distortion)
3:follows)        "gravitational effect" depends on "distance" (in the context of physics)
3:follows)        "gravitational effect" depends on "mass" (in the context of physics)
     and also note "tidal" is generalized by "gravitational effect" (in the context of physics gravity orbit moon satellite distortion)


\end{alltt}
\caption{Knowledge scanned from a distributed application (simulated).\label{story5}}
\end{figure*}

\section{Summary and remarks}

The recent focus of work in artificial intelligence (AI) has been on
the training of artificial neural networks to recognize increasingly
complicated patterns, like voice, faces and handwriting. Trained
networks mimic human cognitive sensory systems, usually one by one.
Neural networks are trained by attempting to replay a lifetime of
experiences at high-speed into a learning process. The extent to which
such a network accumulated, and made predictive, depends on the extent
to which its spatial structure can mimic a representation of the
semantics one happens to be looking for, with only a single
trigger\cite{spacetime3}.  However, it is impossible to measure the
certainty of even this simple kind of reasoning, indicating that an
entirely separate stage of processing by a more deterministic
representation is needed for reasoning that goes beyond a single
inference. That is where the present work fits in. The recognition of
the lexicon of basic patterns is the first part of sensory perception,
the second part is associative reasoning based on its tokens.

In this report, the idea is not to mimic human faculties but to
integrate them, across multiple scales of experience, allowing us to
to add selected semantic, data of high quality, incrementally from any
source to a single model with very few limitations.  The learning
described here is unsupervised, in the AI sense, but still curated by
a surrounding `society' of expert interactions, so it is supervised by
an emergent framework of domain expertise. No learning can happen
without some external selection process, just as no child can be
programmed with life skills without others to raise it. The boundary
between raw exterior and curated interior knowledge is what cognition
enables: to be able to represent and pass on knowledge is a compressed
(tokenized) form for consumption by others, without the need to
undergo every experience, blow by blow, and in person.

These experiments are very interesting, but there are plenty of issues
to address in future work.  Finding how to constrain stories and bring
focus to their trains of thought is the major problem.  With the use
of a spacetime inspired structure, stories are no longer as scarce as
when trying to use an overconstrained typed approach. Lateral
reasoning is now possible, and relevance is maintained by representing
context. The treatment of context in this note is weak, but there is
much room for improvement.  In \cite{lisa98283}, I speculated on the
role of emotional state in determining context.  Emotional weight
plays a significant role in cognition for setting context and
interpretation priorities. From the viewpoint in \cite{spacetime3},
emotions seem to play the role of very coarse aggregate `concepts'
stimulated by sensory/introspective inputs.  Emotional context is thus
a systemic assessment of some `agent' based on a number of contextual
factors. It is a threshold judgement, based on policy:

\begin{center}
\small\sc
\begin{tabular}{c|cc}
         &    positive    & negative \\
\hline
interior &  contented & distressed\\
exterior & like       & dislike \\
prediction& optimistic & pessimistic\\
\end{tabular}
\end{center}
States like distress are easy to measure in a finite system. When a
threshold of behaviour is reached and a system is unable to keep its
promises, e.g. the thrashing to empty a queue. Emotions could be
semanticized versions of these major conditions.

States like good, bad, danger, happy, sad, etc are components of what
we would think of as emotional states. These give clear meanings to
other measurements: how we respond to a context (state) and the
associations we make during good and bad times affect how we recall
concepts later. If we feel strongly about something (good or bad) this
translates into an importance rank of an association.  We want to sum
the recurrence scores of the concepts to label their importance.
It is possible that we might even be able to make all of these by
combining the simultaneous activation of senses with the concepts of
good and bad: e.g.  \beq
\text{Good/bad } \intersect \text{ person } \rightarrow \text{love/hate}\\
\text{Good/bad } \intersect \text{ senses }\rightarrow \text{happy/sad}\\
\text{Good/bad } \intersect \text{ comparison }\rightarrow
\text{true/false} \eeq This remains a topic for future exploration.  I
hope to return to all these issues, in the context of applications, in
future work.

{\bf Acknowledgement:} I am indebted to Steve Pepper for in
depth discussions about linguistics and knowledge representations.  I
am also grateful to Nikos Anerousis, Anup Kalia, Maja Vukovic, and Jin
Xiao for helpful conversations.

\appendix
\section{Example reductions}

\begin{figure*}[ht]
\begin{center}
\includegraphics[width=14cm]{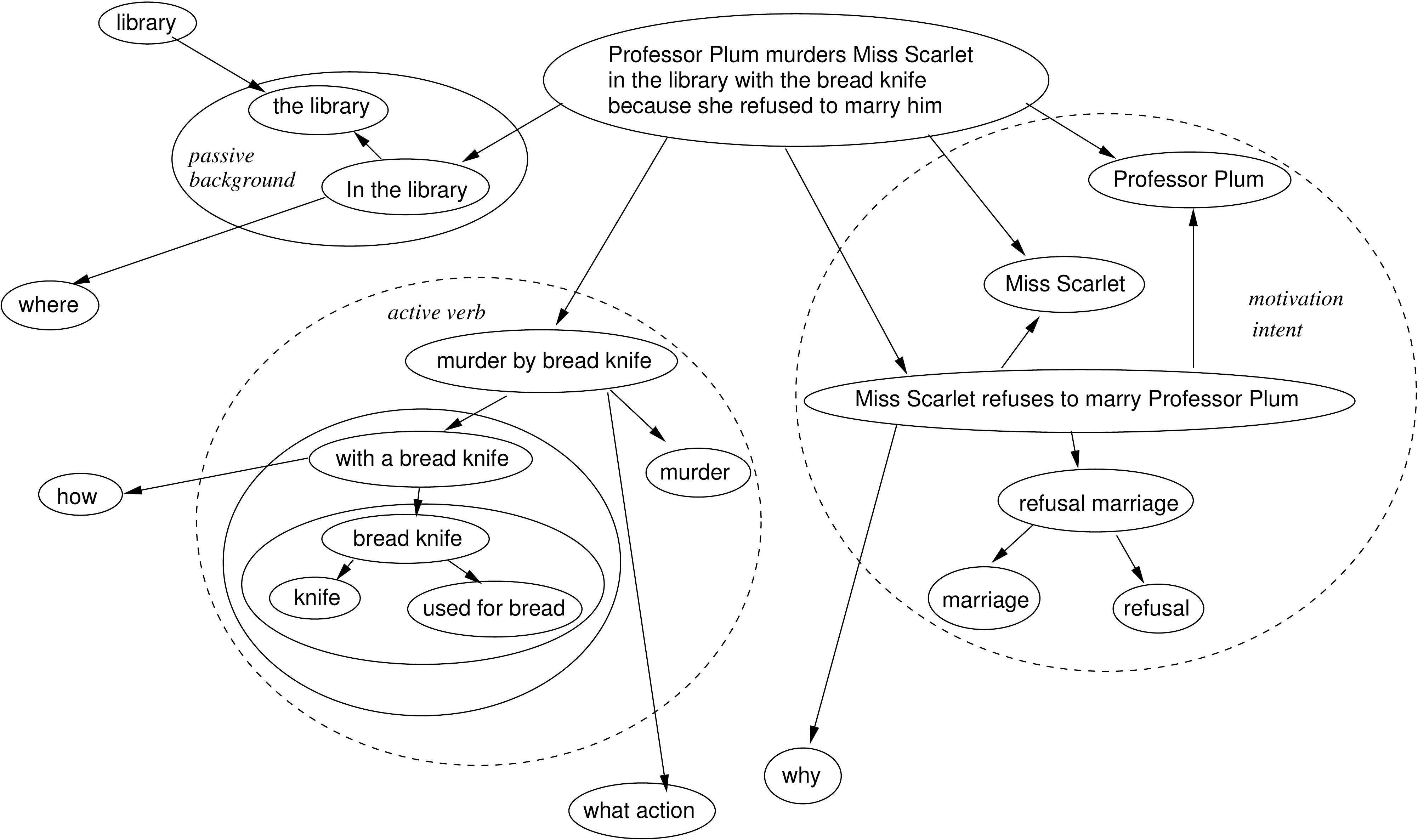}
\caption{\small Representation of the context around the event
  `Professor Plum murders Miss Scarlet with a breadknife in the
  library, because she refused to marry him'. The hierarchy of
  concepts has least invariance at the single proposition, and becomes
  increasingly invariant father from it, towards the edges. Note how
  the context envelops the concepts it includes, and costs much more
  in terms of processing. In compensation, much of its structure is only
temporary, and leaves behind a smaller cheaper residue.\label{cleudo}}
\end{center}
\end{figure*}

\begin{figure*}[ht]
\begin{center}
\includegraphics[width=13cm]{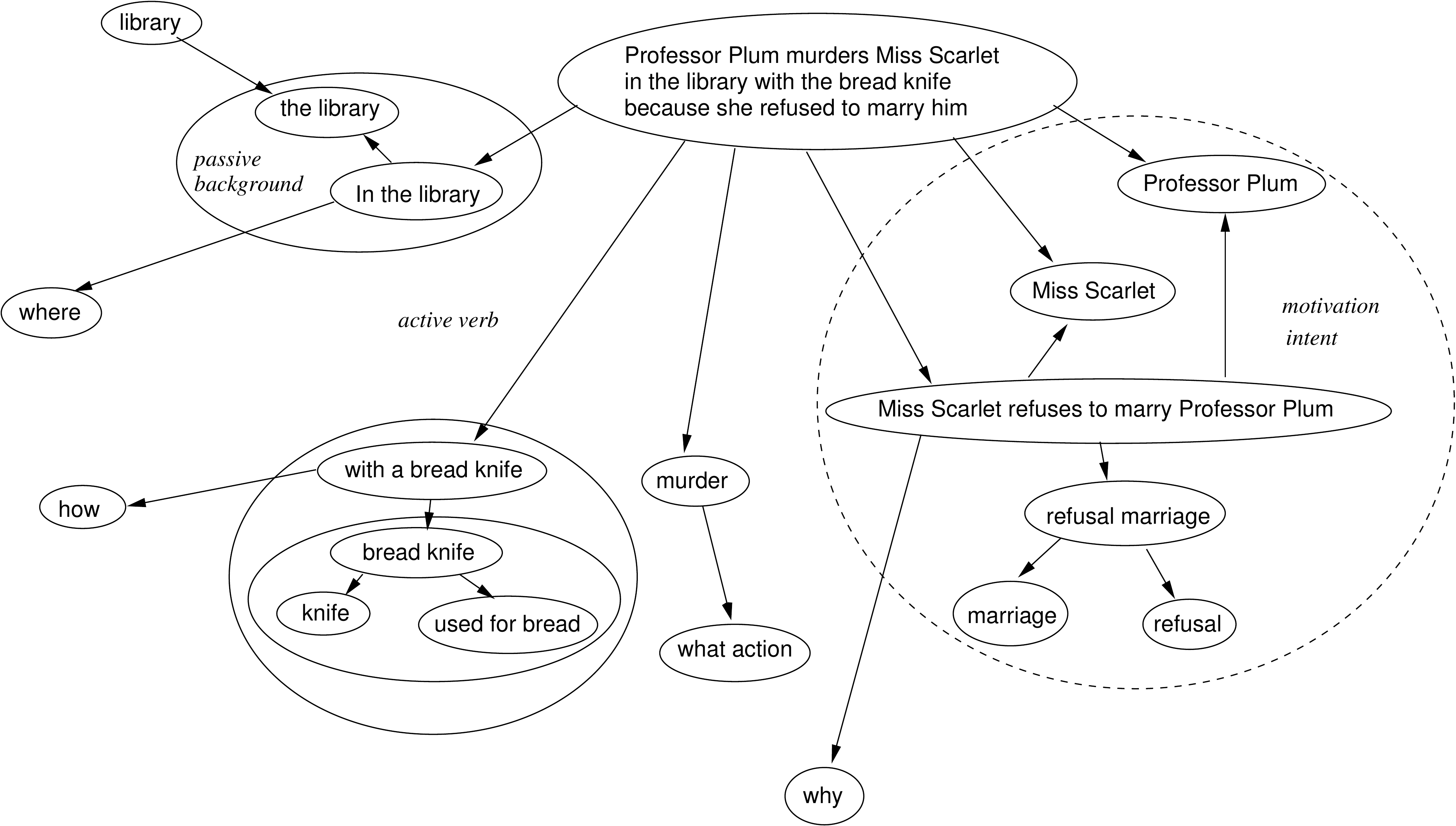}
\caption{\small The example in the previous figure repeated with some of the
levels of hierarchy removed. This looks more like a database structure now, but
loses its ability to generalize subconcepts.\label{cleudo2}}
\end{center}
\end{figure*}

In this appendix, some we examine some examples of the semantic
scaling of concepts, within the hierarchy of increasing invariance.
The following excerpted diagrams help to illustrate different aspects
of the fully cyclic graph of links, picking out particular `parse
trees' from the full structure.  For example, the expression of an
explained event Professor Plum murders Miss Scarlet in the library
with a breadknife, because she won't marry him' is shown in figure
\ref{cleudo}.  The central propositions may be understood as contexts,
composed of a hierarchy of concepts.  Each complex and temporary
proposition is a point of possibly ephemeral permanance, with the
degree of invariance and permanence increasing outwards in rings
around it, reaching a maximum at the most primitive concepts.
Note the intermediate levels of conceptual aggregation in the full
expression.  One could try to arge that these are may be eliminated or
flattened, as in figure \ref{cleudo2}, but this obfuscates the reuse of
concepts, and the ranking of their permanence. 

One might argue that this structure is too difficult to automate, because these
levels of semantics cannot be inferred from the top level proposition; however,
this is thinking upside down. It is the sensors which have to be evolved to
imprint the semantics from the bottom. Propositions based on these semantics
can then make sense in a context formed as (who,what,when,where,how,why).

\begin{figure*}[ht]
\begin{center}
\includegraphics[width=12cm]{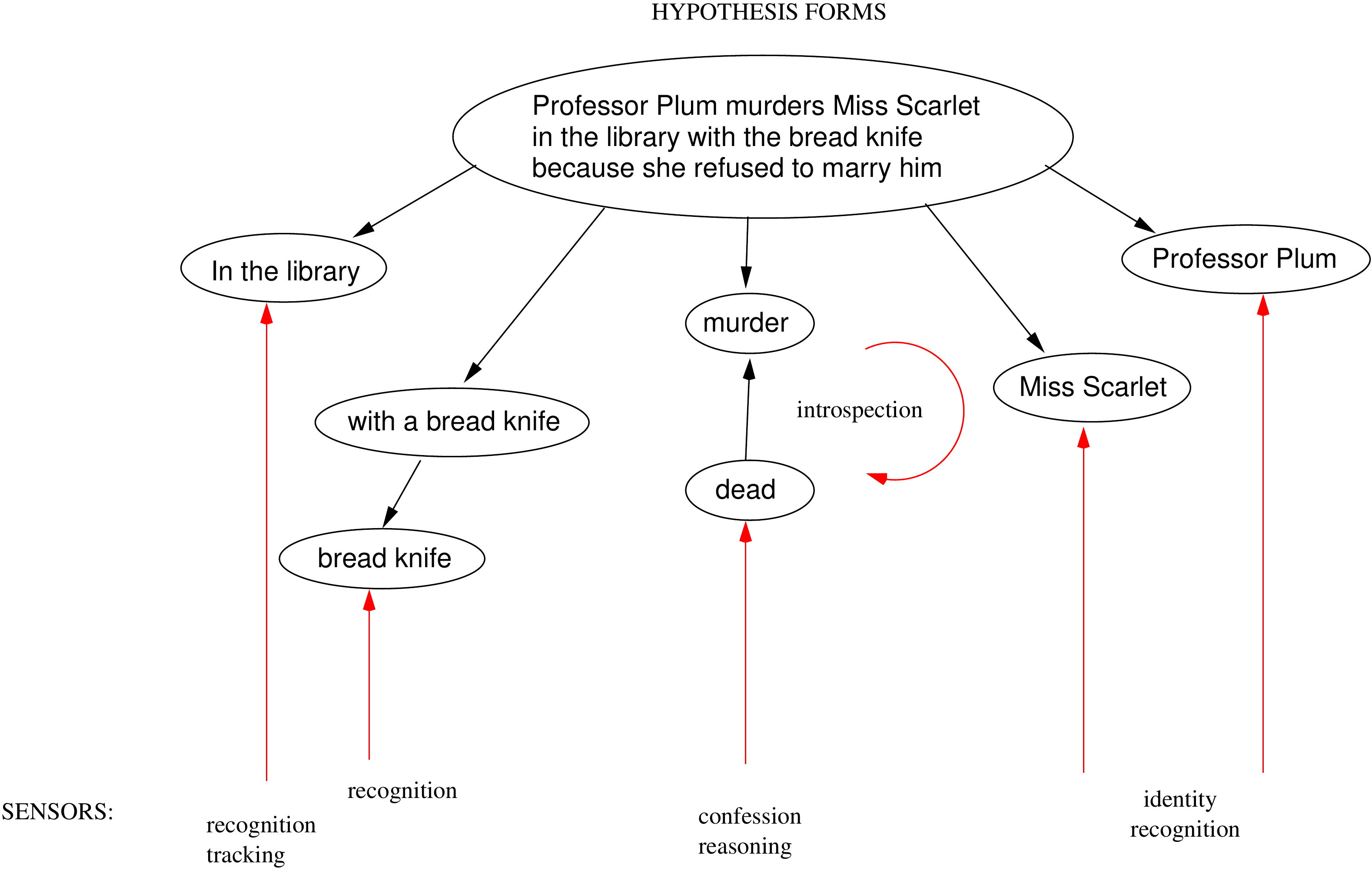}
\caption{\small Smart adapted sensory level events contribute to an aggregate context.\label{cleudo3}}
\end{center}
\end{figure*}

Figure \ref{cleudo3}, shows how sensory inputs and anomalies might
contribute to the formation of the context structure. 
Simply
collecting data from sensors alone will not advance towards a clear
human understanding of the relationships, without a skeleton model on
which to imprint data.  In nature, this might occur by some form of
spontaneous symmetry breaking, and separation of timescales, so that
persistent structures act as the skeleton on which shorter timescale
corrections attract. In this way, weakly non-linear feedback can lead
to emergent structure, but the results will not be humanly comprehensible
unless they map to the particular conceptual structures we have evolved
ourselves. Thus pure emergence is not the right way to bring about human
inspectable results.

\begin{figure*}[ht]
\begin{center}
\includegraphics[width=12cm]{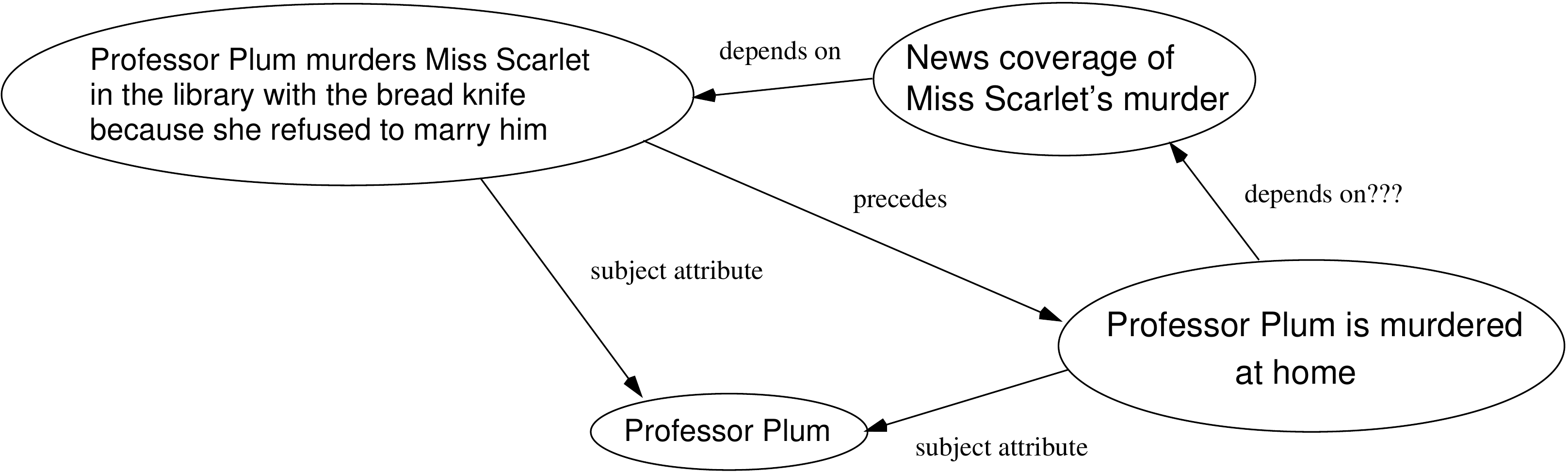}
\caption{\small The large scale structure of models.\label{cluedo4}}
\end{center}
\end{figure*}

From here, if revisited enough times the hypothesis hub will persist
in the `mind' of the cognitive structure and become eligable for connection to other
events and concepts, as in figure \ref{cluedo4}.

From the hierarchy, we can start to see the invariances, and levels of
permanance or timescale sensitivity. Short term context is built from
long-term concepts to create a sense of `here and now'. This is
aggregated into a main structure, which contains fragments of
narrative meaning, over a longer timescale. Then these in turn reduce
to core long-term concepts again. This sequence chases its tail: it is
not an acyclic decomposition, but rather an eigenstate of a cyclic graph.

\begin{figure*}[ht]
\begin{center}
\includegraphics[width=12cm]{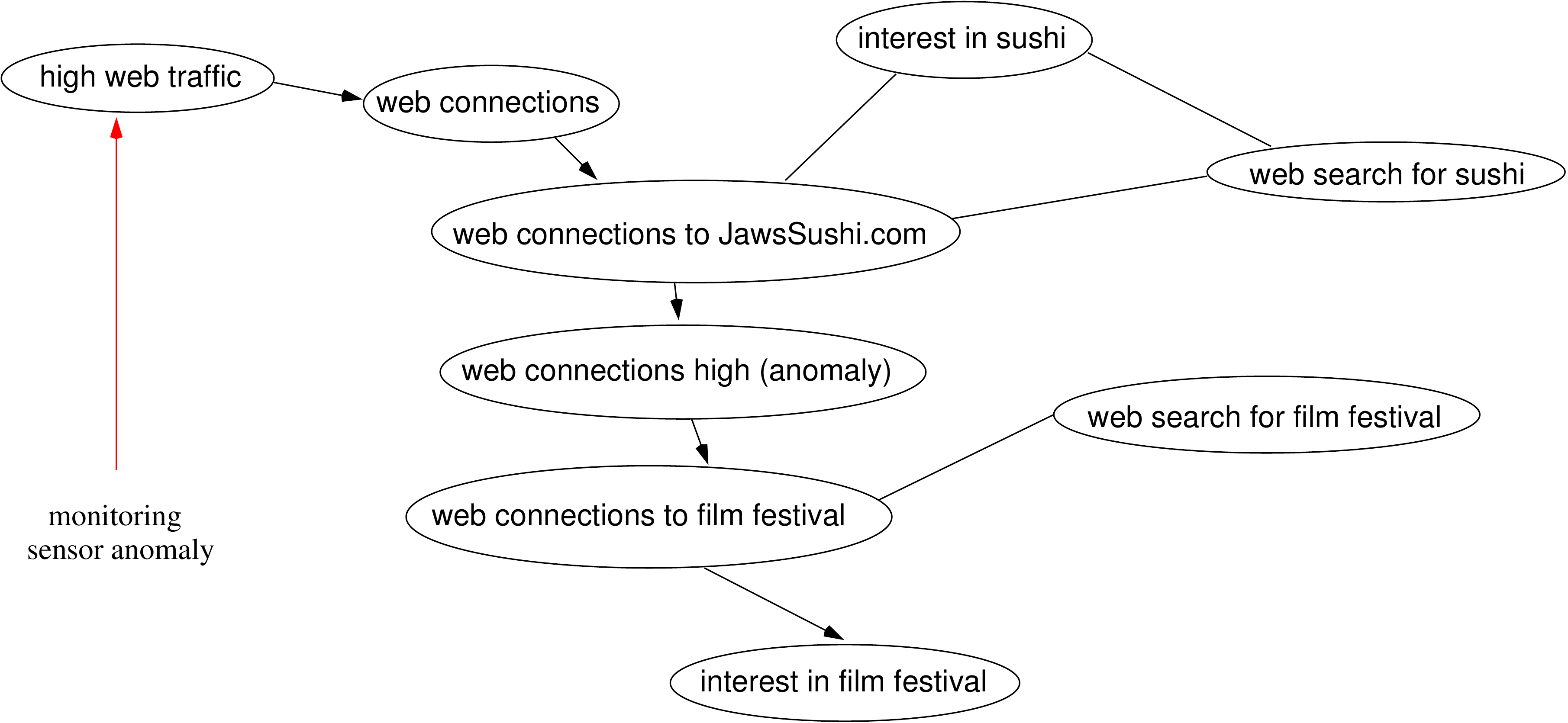}
\caption{\small The large scale structure of models.\label{jaws1}}
\end{center}
\end{figure*}

Consider an example of system measurment on a website, in which
traffic to a sushi restaurant called `Jaws' is high because users are
getting muddled looking for tickets to a film festival featuring the
Steven Spielberg movie `Jaws'. The connection which would allow us to
make an inference is clearly the name jaws, without particular
semantics.
We can postulate what kinds of specialized sensor semaphores we need to
construct this cognitive narrative.
\begin{itemize}
\item Here and now: short term context includes the particular Jaws film festival, web bookings, tickets,
web traffic, etc.
\item Main structure: middle term context includes the existence of the Jaws Sushi restaurant,
the concept of a file festival, interest in a film festival, cinemas, etc.
\item Basis concepts: web connections, festival, film, jaws, sushi, etc.
\end{itemize}
From a high level, if we assume that the sensors and their
aggregations have done their jobs to construct a graph, we obtain a
picture something like figures \ref{jaws1} and \ref{jaws2}.  A causal
parsing based on a web sensor anomaly makes a connection between the
inputs of different sensory semaphores attributing high web traffic on
a trunk line as being due to hits on the website of a sushi
restaurant. This happens to have the same name as a popular movie at a
film festival. Confusion of these websites is the real reason for
anomalous traffic. The brainstorming analysis does not stop on finding the presumed
`root cause', however, it wandering into other territory since it has no reason
to eliminate those pathways. The major task to be solved in future work is the 
selection of only relevant pathways and boundaries.
\begin{figure*}[t]
\footnotesize
\begin{alltt}
\it
./stories -s ``high web traffic''

0:follows)  "high web traffic" may be caused by "web connections" (intended context: online web - 80%)
1:expr-by)    "web connections" is a role fulfilled by "web connections to jawssushi.com" (intended context: browsing the web - 99%)
2:expr-by)      "web connections to jawssushi.com" is a role fulfilled by "high web connections to jawssushi.com" 
                 (intended context: monitoring service performance - 96%)
2:apprxnr)      "web connections to jawssushi.com" may be related to "web connections for jaws" 
                 (intended context: online web - 92%)
3:follows)        "web connections for jaws" may be caused by "interest in film festival" 
                 (intended context: Browsing the web entertainment culture - 80%)
...
\end{alltt}
\caption{\small A causal parsing making a connection between the inputs of different sensory semaphores attributing
high web traffic on a trunk line as being due to hits on a sushi restaurant that happens to have the same name as
a popular movie at a film festival. The analysis does not stop on finding the presumed `root cause', it wandering into other
territory since it has no reason to eliminate those pathways.\label{jaws2}}
\end{figure*}

\bibliographystyle{unsrt}
\bibliography{spacetime}

\end{document}